\documentclass{article}

\usepackage{natbib}

\usepackage{lmodern}
\usepackage{geometry}
\usepackage[utf8]{inputenc} 
\usepackage[T1]{fontenc}    
\usepackage[colorlinks=true, allcolors=black]{hyperref}\usepackage{url}            
\usepackage{booktabs}       
\usepackage{amsfonts}       
\usepackage{nicefrac}       
\usepackage{microtype}      
\usepackage{xcolor}         
\usepackage{subcaption}
\usepackage{amsmath}
\usepackage{amssymb}
\usepackage{mathtools}
\usepackage{amsthm}

\usepackage{paralist}

\usepackage[utf8]{inputenc} 
\usepackage[T1]{fontenc}    
\usepackage{hyperref}       
\usepackage{url}            
\usepackage{booktabs}       
\usepackage{amsfonts}       
\usepackage{nicefrac}       
\usepackage{microtype}      
\usepackage{xcolor}         
\usepackage{amsmath}
\usepackage{graphicx}
\usepackage{booktabs}
\usepackage{algpseudocode}
\usepackage{algorithm}
\usepackage{placeins}
\usepackage{subcaption}

\usepackage{float}
\def\ker{\text{ker}}
\newcommand{\di}{\mathcal{D}}
\newcommand{\mi}{\mathcal{M}}
\definecolor{darkgreen}{rgb}{0,0.75,0.25}
{}

\title{Blind Image Restoration via Merging the Manifold Hypothesis, Kernel Method and a Diffusion-like Flow}

\author{
  \begin{tabular}{c}
    \large Leah Bar$^1$, Liron Mor Yosef$^1$, Shai Zucker$^1$, Neta Shoham$^1$, Inbar Seroussi$^1$, Nir Sochen$^1$ \\[10pt]
    \normalsize $^1$Department of Applied Mathematics, Tel Aviv University, Tel Aviv, Israel \\[5pt]
    \normalsize \texttt{\{lm2, shaizucker\}@mail.tau.ac.il}, \texttt{\{barleah.libra, shohamne\}@gmail.com}, \\
    \normalsize \texttt{\{inbarser, sochen\}@tauex.tau.ac.il}
  \end{tabular}
}

\begin{document}
\maketitle
\begin{abstract}
Blind Image Restoration (BIR) aims to restore degraded images without knowledge of either the type or severity of the degradation. This is a challenging ill-posed inverse problem for which an image prior is essential. We view the prior of an image class as a probability distribution supported on a low-dimensional manifold embedded in the high-dimensional image space. Existing generative approaches primarily model the probability distribution, while the geometric structure is implicit or disregarded. Conversely, geometric approaches do not account for the non-uniform probability distribution on the manifold. We propose a geometric-probabilistic framework for BIR that models these aspects. The probability distribution on the manifold is approximated using a kernel-based method, while a distance from the manifold function induces a geometry-based probability function on the ambient space. This formulation leads to the (Latent) Manifold-Probabilistic Projection Model ((L)MPPM), a restoration model defined in both image and latent spaces. (L)MPPM induces a flow guided by a score field that jointly incorporates geometric and probabilistic information. Evaluated on image restoration tasks and datasets, (L)MPPM consistently outperforms existing approaches, providing empirical evidence for the benefit of incorporating both geometric and probabilistic information into the image prior.
\end{abstract}

\vspace{12pt}

\begin{figure}[ht]
\centering
\includegraphics[width=0.5\linewidth]{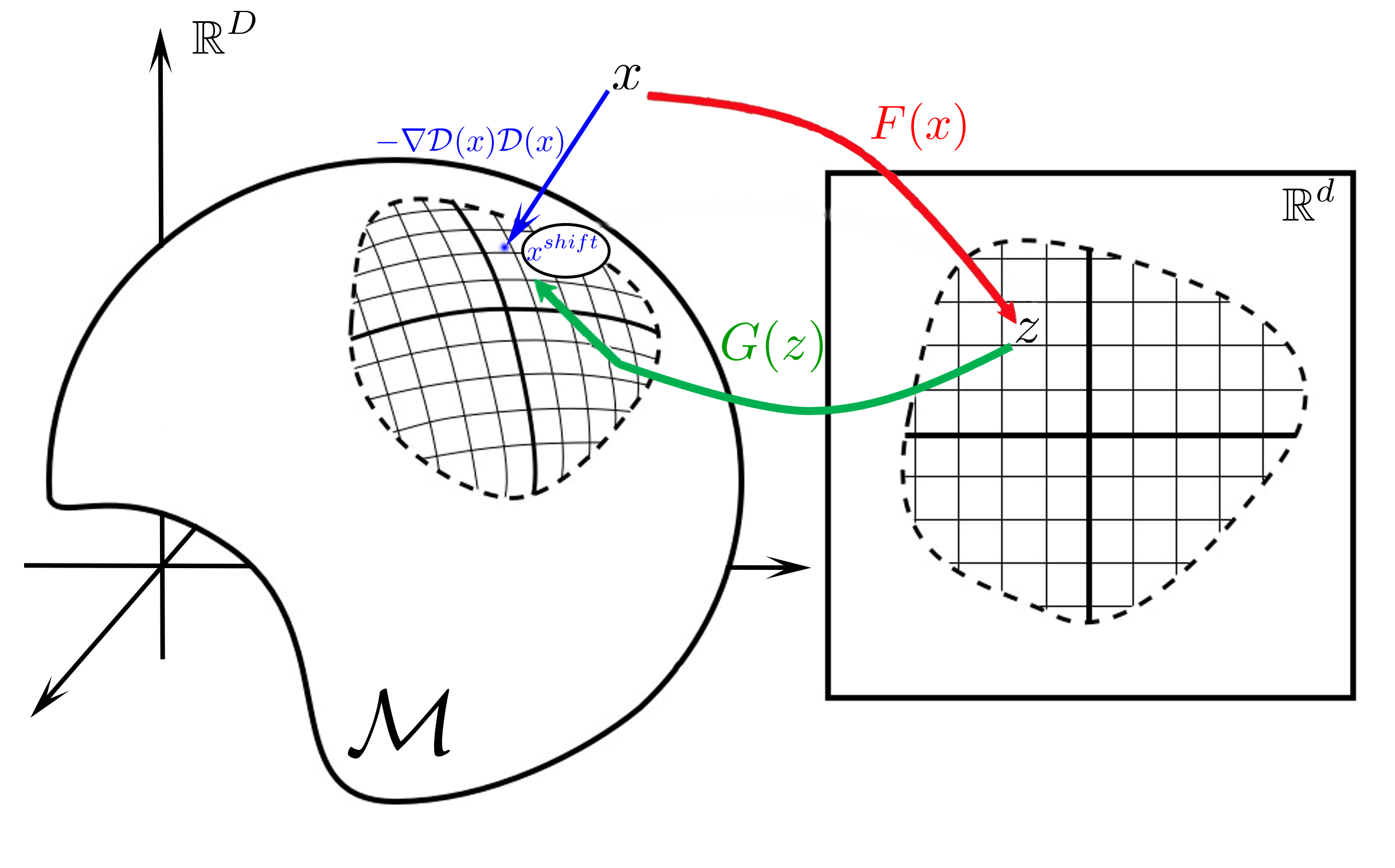} 
\caption{Illustration of our manifold-aware restoration approach. The blue path shows direct projection onto manifold $\mathcal{M}$ using distance function $\di_\mi(x)$, while the red-green path represents encoding-decoding through latent space $\mathbb{R}^d$ via functions $F$ and $G$. Ideally, both paths converge to the same manifold point, ensuring geometrically consistent restoration.}
\label{fig:manifold}
\end{figure}
\newpage
\section{Introduction}

Image restoration aims to recover a clean image from a degraded observation. The problem is inherently ill-posed: a single corrupted image may be compatible with multiple clean images, particularly when the degradation is unknown. Successful restoration therefore requires a prior describing the distribution of plausible clean images. Recent approaches increasingly rely on powerful learned image priors, including diffusion and score-based models, which can guide a corrupted observation toward the distribution of clean images. While these methods have achieved impressive results, they often rely on large neural networks, substantial training datasets, degradation-specific training or modeling, and iterative sampling procedures.

In this work, we propose a unified geometric-probabilistic framework for image restoration based on the assumption that clean images lie near a low-dimensional manifold embedded in the high-dimensional pixel/representation ambient space. Rather than explicitly modeling the degradation, we learn the geometry and probability distribution of the clean-image manifold and use these quantities to guide a corrupted observation toward a plausible clean image. The central object in our formulation is a distance function that assigns to each point in the ambient image space its distance from the clean-image manifold. The gradient of this distance function provides a unit vector field pointing toward the manifold and therefore defines a natural restoration direction.

Our framework combines this geometric description with a probabilistic representation of the clean-image distribution. The generator $G$ parameterizes the clean image manifold. The encoder $F$, on the contrary, operates throughout the ambient space, including points outside the clean-image manifold. We further estimate the probability distribution of clean images in the latent space using a kernel-based method. This distribution is pushed up via the generator $G$ to a distribution over the low-dimensional manifold embedded in the high-dimensional pixel/representation ambient space. These components are combined to define a vector field that guides an off-manifold observation toward a probable point on the clean-image manifold. The resulting process has a diffusion-like interpretation: rather than generating an image from random noise, we start from a degraded observation and iteratively move it toward the clean-image manifold according to the learned geometric and probabilistic structure.

An important distinction from many existing restoration methods is that our framework does not require a model of the degradation. We do not estimate a degradation kernel, learn a degradation-specific network, or require training examples for each degradation type. Instead, the model is trained exclusively using clean images corrupted with Gaussian noise. It is important to note that the Gaussian noise is not added as a model of degradation but as a means to sample the ambient space around the manifold at different distances. At inference time, the same learned model is applied to different additive and non-space-invariant degradations. In our experiments, these include Gaussian noise and structured masking. We emphasize that masking is considered here as an additive corruption where the mask itself (position, form and size) is unknown. Our method does not address the general image inpainting problem in which missing image content must be reconstructed using a known mask and a spatially structured inpainting model.

\subsection{Related Work}

\subsubsection{Image Restoration: Supervised}

Image restoration estimates a clean image from a degraded observation and is inherently ill-posed when the degradation type or its severity is unknown. A large body of work addresses this problem using supervised learning, where networks are trained on pairs of degraded and clean images. These methods can learn highly effective mappings for the degradation distributions represented during training. For example, Restormer uses an efficient Transformer architecture for image restoration tasks including denoising and deblurring \citep{restormer2022}. PromptIR introduces degradation-aware prompts to support all-in-one restoration across several corruption types \citep{promptir2023}. Such approaches demonstrate the effectiveness of discriminative restoration networks, but their training generally requires degraded-clean image pairs and coverage of the degradation distributions that the model is expected to encounter.

Diffusion-based restoration methods instead exploit a learned generative prior and perform restoration through a sequence of denoising operations. 
In image super-resolution, ResShift constructs a residual shift diffusion process \citep{resshift2023}, while SinSR distills a diffusion model into a single-step super-resolution model \citep{sinsr2024}. OSEDiff adapts a pretrained diffusion network for one-step real-world super-resolution \citep{osediff2024}. These approaches illustrate the growing effort to make diffusion-based restoration more efficient while maintaining perceptual quality. 

Our objective differs from that of degradation-specific restoration systems. Rather than learning a direct mapping for a specific restoration task, such as deblurring or super-resolution, we learn the geometry and probability distribution of clean images and use them to guide restoration under different additive corruptions. Importantly, the proposed model is trained using Gaussian noise alone, but at inference time it can restore images corrupted by other types of additive noise.

\subsubsection{Unsupervised and Blind Image Restoration}
When the degradation is unknown, image restoration becomes a blind restoration problem. A growing body of work therefore seeks to restore real-world corrupted images without assuming that the degradation process is directly available. Existing approaches address this challenge in several different ways.

One class of methods relies on degradation simulation. Rather than training only on simple synthetic corruptions, these methods construct more realistic degradation pipelines in an attempt to approximate real-world image formation. For example, SR3~\citep{Saharia2023SR3} has been combined with the second-order degradation model of Real-ESRGAN~\citep{Wang2021RealESRGAN} to improve robustness to realistic degradations. Although such approaches reduce the gap between synthetic and real corruptions, they still require assumptions about the degradation process used to generate the training data.

A second class explicitly estimates the unknown degradation. Blind restoration methods may jointly infer the degradation parameters and the clean image. For example, BlindDPS \citep{chung2023parallel} estimates the unknown blur kernel in parallel with image restoration, while GibbsDDRM \citep{murata2023gibbsddrm} samples the clean image and degradation kernel jointly from their posterior distribution. Such methods provide a principled treatment of unknown degradation, but explicitly modeling the degradation remains an additional component of the restoration problem.

A third class seeks representations that are less sensitive to the particular degradation. DifFace \citep{yue2022difface}, for example, uses a pretrained restoration network to provide a distortion-invariant starting point for diffusion-based restoration. DR2 \citep{wang2023dr2} isolates degradation-invariant low-frequency information through iterative refinement before applying an enhancement module. These methods demonstrate that degradation-invariant representations can improve robustness across different corruptions, but still rely on particular restoration or diffusion mechanisms.

Domain-translation approaches provide another alternative. These methods learn relationships between clean and degraded image domains either by synthesizing realistic training pairs or through unpaired learning. RainDiffusion \citep{wei2023raindiffusion}, for example, uses domain translation for image deraining. Such approaches can reduce the need for paired real-world data, but they still require a formulation of the relationship between the relevant domains.

DiffIR \citep{diffir2023} addresses the computational cost of iterative diffusion restoration by predicting a compact restoration-prior representation. DiffBIR \citep{DiffBIR} explicitly decomposes blind restoration into degradation removal followed by diffusion-prior-based information regeneration, allowing a controllable trade-off between fidelity and realism. SeeSR similarly uses degradation-aware semantic prompts to improve semantic fidelity in real-world super-resolution \citep{seesr2024}. These methods demonstrate the value of powerful pretrained generative priors for blind restoration.

In contrast to these approaches, our method does not explicitly estimate or simulate a degradation model. Rather than learning how a clean image becomes corrupted, we learn the geometry and probability distribution of clean images themselves. The degradation therefore enters only through the observed input. This distinction allows a model trained exclusively with Gaussian noise to be applied to other additive corruptions without introducing a separate degradation-estimation module.

\subsubsection{Manifold-Based Generative Models}

The proposed approach is also closely related to the manifold perspective on high-dimensional data. The manifold hypothesis \citep{review_paper} suggests that natural data occupy a low-dimensional locally smooth structure embedded within a high-dimensional ambient space. This perspective has motivated a broad range of generative models.

Autoencoders (AE), variational autoencoders (VAEs), and generative adversarial networks (GANs) \citep{ae_original, kingma2022vae_1, gan_original} construct mappings from a latent representation of a lower dimension to the high dimension data space. These mappings can be interpreted geometrically as parameterizations of a data manifold. In VAEs and related models, the latent representation additionally provides a probabilistic description of the data.

Other approaches explicitly incorporate particular manifold geometries. Hyperspherical VAEs \citep{davidson2022hypersphericalvae} and hyperbolic VAEs \citep{mathieu2019} adapt latent-variable models to non-Euclidean geometries. Riemannian flow models \citep{gemici2016normalizing, mathieu2020riemannian} incorporate Riemannian metrics into flow-based generative models and explicitly account for geometric properties of the data representation. Normalizing flows \citep{rezende2015variational} learn transformations between a data distribution and a simpler base distribution.

Diffusion and score-based generative models provide a different, more implicit representation of the data geometry. Diffusion models \citep{diffusion_original, ho2020denoising, ho2020denoising} learn a sequence of denoising transformations, while score-based models \citep{song2020} learn the gradient of the log probability density. From a geometric perspective, these approaches can be interpreted as learning vector fields that guide noisy samples toward regions of high data probability.

Our formulation is related to these approaches but differs in explicitly combining three components: a parameterization of the clean-image manifold, a distance function from arbitrary ambient-space points to that manifold, and a probability model on the manifold. The distance function provides a geometric notion of how far an observation is from the set of valid clean images, while the probability model provides information about which regions of that manifold correspond to likely clean images. Their combination yields a restoration vector field rather than a generic generation process.

\subsubsection{Image Generation and Denoising as Projection}
The geometric interpretation of restoration connects our framework to denoising and diffusion-based models. A corrupted image is generally an off-manifold point in the ambient space, and the negative gradient of its distance to the clean-image manifold defines a vector field that moves it toward the manifold. Unlike diffusion and score-based models, which learn vector fields from the data distribution, our vector field is derived explicitly from the learned manifold distance and modulated by the probability distribution on the manifold, combining geometric proximity with probabilistic plausibility.

Our framework extends this construction to the latent space by jointly learning a generator (G), an extended encoder (F), and the manifold distance ($\di_{\mathcal{M}}$). The generator parameterizes the clean-image manifold, while the extended encoder maps arbitrary ambient-space observations to the learned latent representation. These components are coupled through the geometric loss.

Unlike conventional denoising autoencoders \citep{denoising_original_vincent2008extracting}, which learn a direct mapping from corrupted to clean images, our model learns the geometric and probabilistic structure needed to define a restoration flow. This allows it to restore additive corruptions not seen during training. Specifically, Gaussian noise is used during training only to sample points in the ambient space, while the learned manifold distance and probability structure guide restoration from other additive corruptions at inference time.

\subsection{Key Contributions}
The suggested formulation provides a unified treatment of multiple additive non-convolution degradations without explicitly identifying the degradation type or its magnitude. In particular, the restoration procedure does not require an externally specified number of diffusion or restoration steps. The flow proceeds according to the learned geometric and probabilistic quantities and can be stopped according to the convergence criterion that depends on the distance from the manifold and/or the high probability as described in Section~\ref{sec:methods}. Thus, unlike approaches in which the sampling trajectory is explicitly parameterized by a diffusion time or noise level, our restoration process is not tied to a predefined degradation schedule.

Another important characteristic of our approach is its relatively small dataset and computational requirements. The model is trained using only a few thousand clean training images, rather than the very large datasets commonly used to train modern generative models. Furthermore, the networks used by our method are lightweight, with a total model size of approximately 32 MB. This makes the proposed framework substantially smaller than many diffusion-based restoration systems while retaining the ability to generalize across additive degradation types.

The geometric interpretation of the proposed approach is related to the manifold hypothesis \citep{review_paper}, according to which high-dimensional data often concentrate near a lower-dimensional manifold. Autoencoders (AE), variational autoencoders (VAEs), and generative adversarial networks (GANs) \citep{ae_original, kingma2022vae_1, gan_original} provide different ways of representing such a manifold through a mapping from a latent space to the image space. Our formulation extends this perspective by explicitly introducing a distance function to the manifold and using it to define a restoration vector field in the ambient space. In contrast to conventional denoising autoencoders, whose objective is typically to map corrupted samples directly to clean samples, the proposed formulation explicitly represents the geometry of the clean-image manifold and the probability distribution along it.

The resulting framework establishes a connection between geometric projection and diffusion-like restoration. A corrupted image can be regarded as an off-manifold point, and the learned distance gradient provides a direction toward the nearest point on the manifold. The probability model also distinguishes between different regions of the manifold, allowing the restoration trajectory to favor probable clean images rather than performing merely geometric projection. This provides a unified interpretation of distance-based projection, probabilistic modeling, and diffusion-like flows within a single restoration framework.

We evaluate the proposed approach on MNIST, SCUT-FBP5500, and CelebA-HQ-256. The model is trained only with additive Gaussian noise as an ambient space sampling method and evaluated on several additive non-convolutional degradation types, including degradations not encountered during training. We compare our method with representative image-restoration approaches, including Latent Diffusion Models (LDM) \citep{rombach2022ldm} and DiffBIR \citep{DiffBIR}. The experiments demonstrate that the learned clean-image geometry can provide effective restoration across multiple additive degradations while requiring only a few thousand training samples and a lightweight model.

The main contributions of this work are:
\begin{itemize}
\item In section~\ref{sec:methods} we introduce a unified geometric-probabilistic framework for image restoration based on an explicit distance function to the clean-image manifold and a probability model defined on that manifold.
\item We demonstrate restoration across multiple additive degradation types without explicitly modeling, estimating, or identifying the degradation in section~\ref{sec:exper}. The model is trained exclusively with Gaussian noise.
\item The proposed restoration flow does not require a predefined degradation-dependent number of restoration steps and instead follows the learned geometric and probabilistic structure.
\item We demonstrate that the proposed approach can be trained using only a few thousand samples and a lightweight network of approximately 32 MB.
\end{itemize}
Overall, the framework connects manifold learning, probabilistic modeling, and diffusion-like restoration, using the geometry of the clean-image space as a degradation-independent prior rather than explicitly modeling the corruption process.

\section{Background and Theoretical Framework}

\subsection{Geometry: Manifold Hypothesis - Explicit}
Many generative networks assume that images lie on a lower-dimensional manifold  defined according to the latent space representation, which is embedded within a higher-dimensional representation space, such as the pixel space or ambient space. This manifold is explicitly modeled by the decoder in autoencoders (AEs) and variational autoencoders (VAEs), and by the generator in various Generative Adversarial Network (GAN) architectures. In all of these models, the manifold $\mi$ is represented as a Monge patch. Let the latent space be $d$-dimensional, parameterized by $z$, and the pixel space be $D$-dimensional, parameterized by $x$, that is (see Fig.~\ref{fig:manifold}):
\begin{equation*}
    G(z)=\big(x_1(z_1,\ldots,z_d),\ldots,x_D(z_1,\ldots,z_d)\big).
\end{equation*}
In simple terms, the value at each pixel in the image (or in similar manifold-structured data) is a function of the $d$ parameters $z$. In this work, we use an autoencoder (or Denoising AE) where the encoder $F(x)$ maps an image $x\in\mathbb{R}^D$ to a point $z\in\mathbb{R}^d$. The generator $G$ is as above and is used as the map that defines the manifold. 

 Many works, in the context of deep learning, use this representation to analyze the dataset as a Riemannian manifold.  We will mention here, as examples, \citep{Shao2018,Wang2021} where geodesics and directions of meaningful changes on the manifold are studied. In \citep{chadebec2022geometric} the relation of the induced metric of the manifold was found to be approximated close enough to the encoded point of a clean image by the inverse covariance found in VAE.

\subsection{Geometry: Distance Function - Implicit}
Another (implicit) way to describe a manifold is as the zero level set of a function. The distance function to the manifold in the ambient (representation) space is well suited for this purpose and is defined as follows:
\begin{equation}
\di_\mi(x)=\min_{y\in \mi}\|x-y\|,
\end{equation}
where $\|\cdot\|$ denotes the Euclidean norm. In this high-dimensional representation space, the distance function provides a natural measure of the proximity of a point to the manifold. It is well known that $\di_\mi$ satisfies the Eikonal equation \citep{Hamilton1828}
 $\|\nabla \di_\mi(x)\|= 1$, with the natural boundary condition $\di_\mi(x)=0$ for all $x\in \mi$. Moreover, it is clear that $-\nabla \di_\mi(x)$ defines a unit vector field pointing in the direction of the shortest path to the manifold.

\subsection{Probability: Ambient Space} 
On the probabilistic side, we follow works such as \citep{Kadkhodaie-etal2023} and \citep{Sun-etal2025}. We assume that the clean data are sampled from an underlying distribution, denoted by $P_c$. Our goal is not to recover this distribution exactly. Rather, we construct a probability model whose score provides a useful vector field for guiding degraded observations toward the clean-data manifold and toward its more densely populated regions.

This objective can be achieved by defining a probability function that decreases monotonically with the distance from the manifold. More precisely, the probability assigned to a non-data point $x$ is defined so that the resulting score vector field points toward the data manifold and its more densely populated regions. To this end, we naturally define the conditional probability of a corrupted image $x$ given a clean image $x'$ as
$$
P(x \mid x') = f\bigl(\Delta(x,x')\bigr),
$$
where $\Delta(x,x')$ denotes the distance between the clean and corrupted images, and $f$ is a monotonically decreasing function.

For ease of analysis and computation, we choose $f$ to be Gaussian. This choice is not meant to suggest that the Gaussian is the “true” underlying function, but rather that it is sufficient for producing the desired vector field. As a second standard assumption, we take $\Delta(x,x') = {\|x - x'\|}$ to be the Euclidean distance. These considerations lead to the following expression for the conditional probability:
\begin{equation}\label{cond-prob}
P_{\sigma}(x \mid x') = \frac{1}{Q_\sigma}\exp\left(-\frac{{\|x - x'\|}^2}{2\sigma^2}\right),
\end{equation}
where $Q_\sigma$ is the appropriate normalization constant.

The resulting probability distribution over the ambient space is then given by
the marginal of the joint probability of a point in image space and a point from the distribution of real clean images:
\begin{equation}\label{eq:prob-ambient_marginal}
P(x) = \int_{\mathbb{R}^D} P(x \mid x') P_c(x')\, dx'
\end{equation}
where $P_c(x')$ denotes the probability density of clean images. Next we use our assumption about the conditional probability and use $P_\sigma(x)$ as the conditional probability
\begin{equation}\label{eq:prob-ambient}
P(x;\sigma) = \int_{\mathbb{R}^D} P_{\sigma}(x \mid x') P_c(x')\, dx'
= \frac{1}{Q_\sigma} \int_{\mathbb{R}^D} \exp\left(-\frac{{\|x - x'\|}^2}{2\sigma^2}\right) P_c(x')\, dx',
\end{equation}
which is the well-known expression from \citep{Kadkhodaie-etal2023, Sun-etal2025}:

Although the Gaussian form may appear to impose a restriction to Gaussian noise, this formulation makes no explicit assumption about the type of degradation process that maps $x'$ to $x$. It only postulates that the probability of the degraded image $x$ to be originated from the clean image $x'$ declines exponentially with the distance between the images. 
As a result, this approach yields a blind image restoration framework that does not require prior knowledge of the degradation type or its magnitude.  
However, computing this probability distribution is intractable in practice, as noted by \citep{Kadkhodaie-etal2023, Sun-etal2025}, since it requires evaluating an expectation over the entire ambient space.

\subsection{Probability and Geometry: The Manifold Hypothesis} 
We provide an additional formulation of the probability distribution based on the manifold hypotheses.  Namely, we propose that the probability of clean images is concentrated on a low-dimensional manifold

\begin{equation}
    P_c(x')=\int P(G(z))\delta(x'-G(z))dV_\mi,
\end{equation}
where $z=F(x')$ is the latent representation of the clean image $x'$, and $\delta$ denotes the Dirac delta function and $dV_\mi$ is the volume element of the manifold.
This means that $P_c(x')=P(G(z))$ per manifold's volume unit if $x'\in\mi$ and $0$ otherwise. Substituting $P_c(x')$ in~\eqref{eq:prob-ambient} and after changing the order of integration, we obtain (the derivation is given in Appendix~\ref{app:non_uniform}):
\begin{equation}
\label{eq:kernel}
    \begin{aligned}
   {P_{\text{non-u}}(x;\sigma)} &= \int_{\mathbb{R}^d} P_\sigma(x\mid G(z))P(z)dz.
    \end{aligned}
\end{equation}
{Note that $P_{\text{non-u}}(x;\sigma)$ accounts for the nonuniform distribution of clean images on the manifold, and the integration is performed in the \emph{latent space}.} In the limit $\sigma\to 0$, only the point on the manifold closest to $x$ contributes significantly, and for small enough $\sigma$, denoted $\sigma_d$, we obtain:
\begin{equation}
\label{distance-prob}
P_{\sigma_d}(x\mid G(z))\propto P_d({x}) {:=}\frac{1}{Q_d}\exp\left({-\frac{\di^2_\mi({x})}{2\sigma_d^2}}\right),
\end{equation}
where $Q_d$ is a normalization factor. In this limit, we obtain an ``Energy-based model'' (EBM) where $E=\di_\mi^2$. We can ignore the limiting process by which this EBM was derived and postulate it with a variance $\sigma_d^2$ of our choice. 
It is worth noting that learning the distance function $\di_\mi$ to the manifold directly makes the algorithm free of time or noise-level conditioning. Unlike diffusion models, which require the noise level (or time step) as an input and a predetermined number of sampling steps, our method simply moves toward the manifold. It therefore needs no prior knowledge of the number of steps or of the degradation level, and the iterations can be stopped once the manifold is reached. This is because the learned distance encapsulates the noise level or time step through the actual quantity of interest: the distance to the manifold of clean images. In this way, it resolves another challenge associated with using diffusion models for image restoration \citep{Sun-etal2025}.

\subsection{Kernel Density Estimation in Latent Space}
{In the formulation of $P_{\text{non-u}}(x;\sigma)$~\eqref{eq:kernel},\eqref{distance-prob}}, the probability at $x$ is obtained by integrating contributions from all points on the manifold, where the conditional probability depends solely on the distance to the manifold and is thus purely geometric. Each contribution is weighted by $P(z)$, which represents the likelihood that the point $G(z)$ on the manifold corresponds to a clean image. 

Since the distribution $P(z)$ is unknown, we estimate it using a kernel density method \citep{Rosenblatt1956, Parzen1962}, a.k.a. ideal denoiser with delta mixture distribution/empirical distribution \citep{Wang-etal2024, Karras-etal2020}:
\begin{equation}
P(z)\approx {P}_{\text{ker}}(z) := \frac{1}{Q_{\text{ker}}}\sum_{\alpha\in S} \exp\left({-\frac{\|z-z_\alpha\|^2}{2\sigma_{\text{ker}}^2}}\right),
\label{eq:pker}
\end{equation}
where $S$ is the set of latent code indices corresponding to clean images, and {$Q_{\text{ker}}= |S|(2\pi\sigma_{ker}^2)^{d/2}$}  is the normalization constant. 

\begin{figure}[h]
\centering
\includegraphics[width=0.5\linewidth]{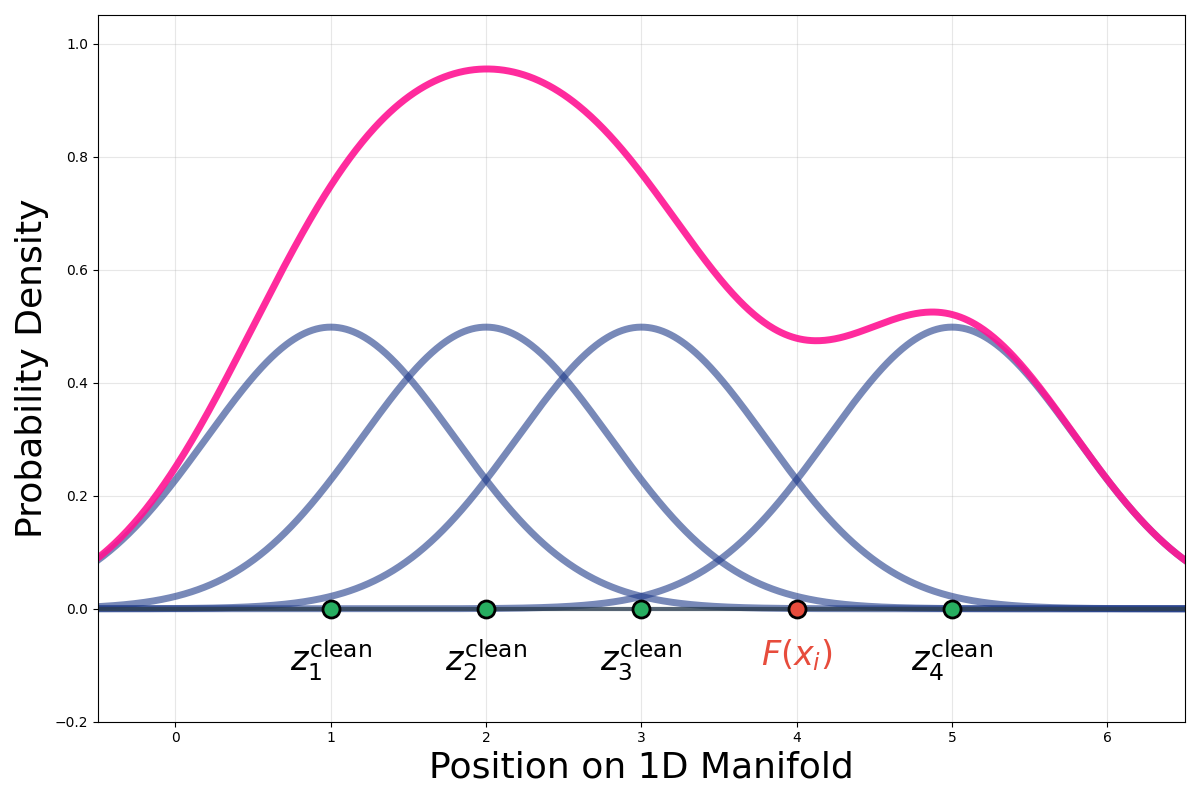}
\caption{An illustration of the kernel approximation $P_{\text{ker}}(z)$ of the probability distribution  $P(z)$ in the latent space. }
\label{fig:rbf-latent-space}
\end{figure}

In Fig.~\ref{fig:rbf-latent-space}, we illustrate $P_{\text{ker}}(z)$. Clearly, the encoding of a generic image $x$ in the latent space, i.e., $F(x)$, may lie in a region with low probability. The probability of a point $x$ being an image depends on its distance to every point on the manifold, weighted by the probability of that point in the latent space. {Using this kernel approximation together with the conditional probability~\eqref{distance-prob}, Eq.~\eqref{eq:kernel} takes the form }  
\begin{equation}
    P_{\text{non-u}}(x;{\sigma})\approx {\hat{P}_{\text{non-u}}(x;{\sigma}):=}\frac{1}{Q_{\sigma} Q_{\text{ker}}}\sum_{\alpha\in S}\int_{\mathbb{R}^d} \exp\left({-\frac{\|{x}-G(z)\|^2}{2{\sigma}^2}}\right)\exp\left({-\frac{\|z-z_\alpha\|^2}{2\sigma_{\text{ker}}^2}}\right)dz.
    \label{eq:pnonu}
\end{equation}

\section{Geometric Perspective}\label{geometric-perrspective}
Since both the encoder $F$ 
and the manifold distance $\di_\mi$ are defined over the ambient space $\mathbb{R}^D$, training mappings that induce a diffusion-like flow from corrupted images back to clean manifold samples requires sampling in this high-dimensional space. However, this sampling is inherently difficult due to the curse of dimensionality. Following diffusion models, we obtain ambient samples by adding \emph{high-variance Gaussian noise} to the data points. Although this strategy does not cover all possible corruptions, it empirically yields effective mappings. Notably, despite being trained solely with Gaussian noise, our models generalize well to other types of image corruption at test time, (Figs. \ref{fig:patches_0.08},\ref{fig:scribble_5},\ref{fig:nose_18},\ref{fig:left_eye_18}).

\begin{figure}[H]
\centering
\includegraphics[width=0.7\linewidth]{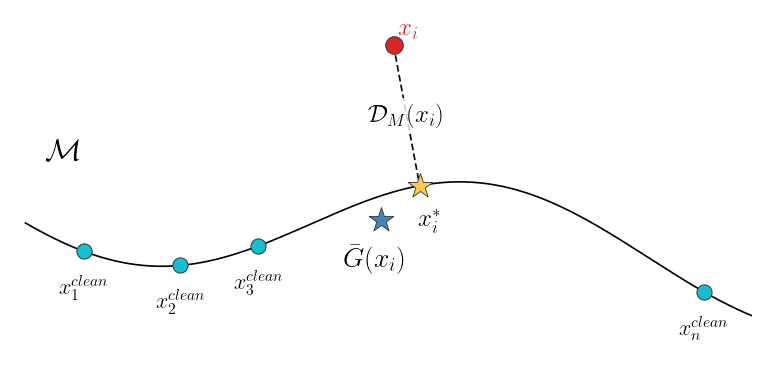}
\vspace{-10pt}
\caption{The manifold $\mathcal{M}$ is illustrated as the curved line. $x_i^*$ is the closest point to $x$ on the manifold. $\bar{G}(x)$ is depicted as well and is not necessarily a point on the manifold.}
\label{fig:x-star}
\end{figure}

\begin{figure}[h]
\centering
\includegraphics[width=0.93\linewidth]{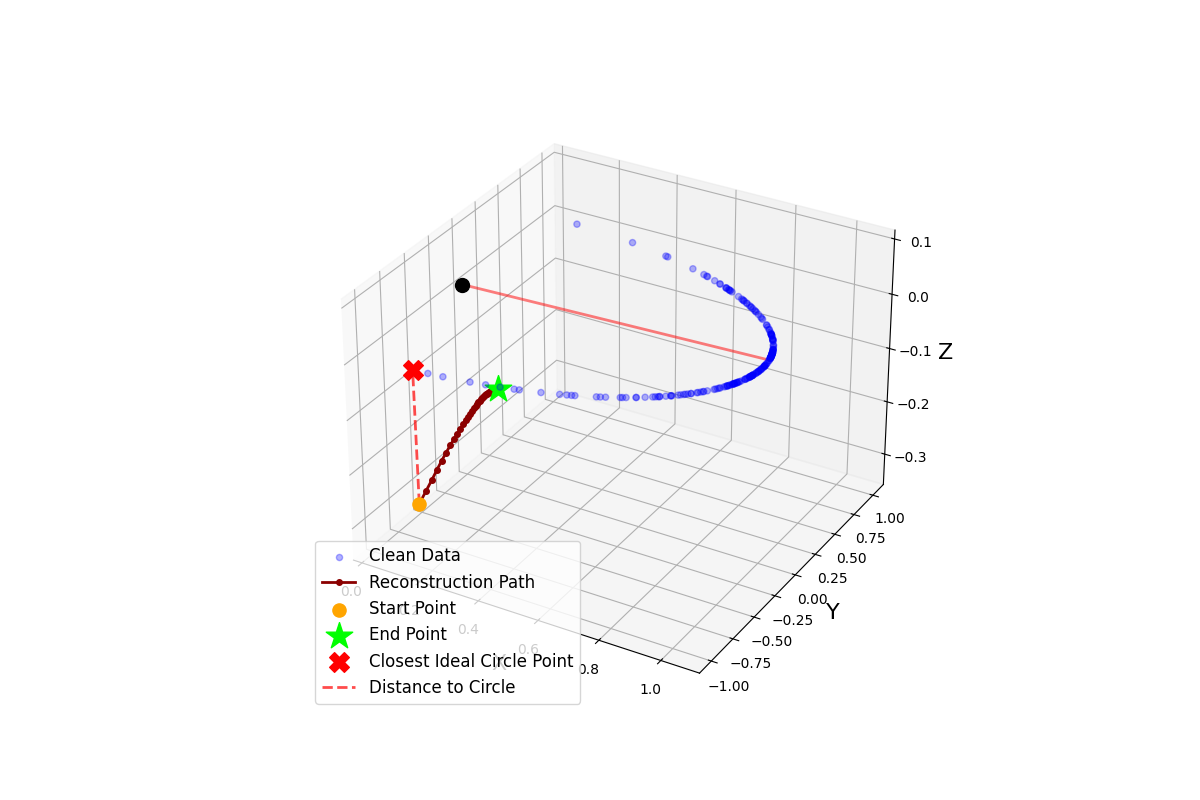}
\caption{The manifold $\mathcal{M}$ is the unit circle lying in the $\text{xy}$-plane and is parametrized by the azimuth angle $\theta$. It is sampled according to a normal distribution centered at $\theta_0$ indicated by the red line. The reconstruction trajectory is shown in dark red. Note that the final result of the iterations on $x$ does not converge to $x^*$ which is the closest point on the circle. Instead, it is influenced by the data distribution on the manifold through the effect of $\bar{G}(x)$.}
\label{fig:circle-illustration}
\end{figure}

To relate a corrupted image to its clean projection, we use the concept of the \emph{score}. The score is a $D$-dimensional vector field defined by $s(x) = \nabla_x \log P(x)$, which points in the direction of the steepest ascent of the probability density. For the distance-based probability distribution $P_d(x)$~\eqref{distance-prob}, we obtain:  
\begin{equation}\label{eq:s_d}
s_d(x) = \nabla_x \log P_d(x) = \frac{\nabla_x P_d(x)}{P_d(x)} = -\frac{1}{\sigma_d^2}\di_\mi(x)\nabla_x \di_\mi(x). 
\end{equation}
Since $\di_\mathcal{M}(x)$ is the distance to the manifold, its gradient $\nabla_x \mathcal{D}_\mathcal{M}(x)$ is a unit vector pointing to the closest point on the manifold. Therefore, for $\sigma_d=1$ we have: 
\begin{equation}\label{eq:x_shift}
x^{\text{shift}} := x+s_d(x) = x - \di_\mathcal{M}(x)\nabla_x \di_\mathcal{M}(x) =G(F(x))=x^*, 
\end{equation}
where $x^{*}$ is the point on the manifold closest to $x$ (see Fig.~\ref{fig:x-star}).\\ 
To incorporate the probability distribution of clean images on the manifold (or equivalently, in the latent space), we interpret the probability in the ambient space as a marginal distribution. This allows  the approximation of the score function using a kernel-based method: 
$$
s(x)=\nabla_x \log {P}{_{\text{non-u}}(x)}\approx \nabla_x\log \hat{P}{_{\text{non-u}}(x)} =:\hat{s}(x).
$$
Direct computation results in 
\begin{equation}\label{eq:s_k_hat}
\hat{s}(x) = -\frac{1}{2\sigma^2}\left( x-\bar{G}(x)\right), 
\end{equation}
where $\bar{G}(x)=\sum_{\alpha\in S} \bar{G}_\alpha(x)$, and
\begin{equation}
\bar{G}_\alpha(x) =  {\frac{1}{\hat{P}_{\text{non-u}}(x)Q_{\ker}}}\int\left[ G(z) P(x\mid G(z))\exp\left({-\frac{\|z-z_\alpha\|^2}{2\sigma_{\text{ker}}^2}}\right)\right]dz.
\label{eq:G_bar_alpha}
\end{equation}   
Note that $\bar{G}(x)$, which is the mean of the contributions from all points on the manifold to the probability $P{_d}(x)$ 
does not necessarily lie on the manifold. In contrast, $x^* = G(F(x))$ is, by definition, a point on the manifold. See Fig.~\ref{fig:x-star} for an illustration and Fig.~\ref{fig:circle-illustration} for a synthetic example.

Compared to prior work, $\bar{G}(x)$ in Eq.~(\ref{eq:s_k_hat}) is the same object as $\mathbb{E}_{y\in P_c}[y|x]$ in  \citep{Kadkhodaie-etal2023}. This quantity is intractable because the integral can't be approximated by sampling clean images beyond the points in the dataset. It is therefore often replaced by the denoiser $x^*$ \citep{Kadkhodaie-etal2023}.  Clearly, the approximation of $\bar{G}$ by $x^*$ is justified only under a uniform distribution over the manifold. In contrast, in our geometric formulation the integral over $z$ in the computation of $\bar{G}_\alpha(x)$  is approximated by randomly sampling the normal distribution centered around the training point $z_\alpha$ (for details, see Appendix~\ref{app:score}.). 

A noisy or corrupted image $x$ can be viewed as a point in the ambient space. The image generation then becomes the task of finding an appropriate, though not necessarily orthogonal, projection of this point onto the manifold of clean, meaningful images. If the mappings and functions $G$, $F$, and $\di_\mi$ are perfectly accurate, a single step can move $x$ closer to the corresponding clean image. Since the ambient space is sampled sparsely, particularly in regions far from the manifold, the learned approximations become less accurate as the distance from the manifold increases. To mitigate this, we apply multiple iterative steps that progressively refine the estimate as the trajectory approaches the manifold. This procedure resembles a diffusion-like flow. 

Equations~\eqref{eq:s_d} and~\eqref{eq:x_shift} motivate a diffusion-like process guided by the distance function. The score induces a vector field in the ambient space. Using the Tweedie formula \citep{Efron2011}, we update the sample by taking a step toward the manifold projection, i.e., toward the closest point on the manifold. This yields the iterative update
\begin{equation}\label{eq:x_follow_D}
   x^{n+1}=x^n - \alpha \di_\mi(x^n)\hat{\nabla}_x\di_\mi(x^n)\quad \text{\rm with} \quad 0<\alpha<1\quad \text{\rm and}\quad x^0={x}
\end{equation}
{where
\begin{equation}
\hat{\nabla}_x \di_\mi(x) := \frac{\nabla_x \di_\mi(x)}{\|\nabla_x \di_\mi(x)\|}
\label{eq:normed_grad}
\end{equation}
denotes the normalized gradient of the distance function. Since the distance network provides only an approximation, we normalize the gradient to better control the step size.} Moreover, it can be easily shown (see Appendix~\ref{app:tweedie}) that this flow converges to the manifold since 
\begin{equation}
    \di_\mi(x^{n+1})=(1-\alpha)\di_\mi(x^n) + O(\alpha^3).
\end{equation}
Equation~\eqref{eq:x_follow_D} does not take into account the distribution of training samples along the manifold. To address this limitation, we incorporate the score induced by the kernel formulation and, by applying the Tweedie formula (see Appendix~\ref{app:tweedie}), to obtain the following score:
\begin{equation} 
\label{eq:x_follow_D_G_avg}
x^{n+1}=(1-\beta)x^n + \beta \bar{G}({x^n}) - \alpha \di_\mi(x^n)\hat{\nabla}_x\di\mi(x^n),
\quad \text{with } 0<\alpha,\beta,\ \alpha+\beta<1,\  x^0=x.
\end{equation}
The trajectory of $x$ as it moves towards the manifold is illustrated in Fig.~\ref{fig:circle-illustration}.
\section{Methods}
\label{sec:methods}
\subsection{Manifold-Probabilistic Projection Model (MPPM)}
The autoencoder and the distance function are implemented as separate neural networks and are jointly trained using the following loss function 
\begin{equation}
    \begin{aligned}
        \mathcal{L}(F,G,\di_\mi) 
        &= \lambda_1\sum_{x\notin \mi} \left(\di_\mi(x) - \|x - x^{*} \|\right])^2+\lambda_2\sum_{x_i\in \mi} \left\|x^{\text{clean}}-G(F(x^{\text{clean}}))\right\|^2\\
        &+\lambda_3\sum_{x\in \mi} |\di_\mi(x)|^2
        +\lambda_4\sum_{x\in\mathbb{R}^D}\rVert x^{\text{shift}}-x^*\rVert^2,
    \end{aligned}
    \label{eq:main_loss_ambient}
\end{equation}
{where 
\begin{equation}
x^{\text{shift}}=x - \di_\mi(x)\hat{\nabla}_x\di_\mi(x)
\end{equation}
and
\begin{equation}
x^*=G(F(x)).
\end{equation}
}
 The first term in \eqref{eq:main_loss_ambient} defines the distance function assuming a perfect autoencoder; the second is the standard autoencoder loss. The third term enforces the boundary condition on the distance function. The last term enforces the geometric consistency of Eq.~(\ref{eq:x_shift}) (see also Fig.~\ref{fig:manifold}). {At inference, we use the Tweedie formula~\eqref{eq:x_follow_D_G_avg}, where
\begin{equation}
\bar{G}_\alpha(x)\approx \frac{\sum_{z\in \mathcal{N}(z_\alpha,\sigma_{\ker}^2)}
    G(z)\exp\left(-{\|{x}-G(z)\|^2/2\sigma^2}\right)}{\sum_{z\in \mathcal{N}(z_\alpha,\sigma_{\ker}^2)}
    \exp\left({-\|{x}-G(z)\|^2/2\sigma^2}\right)}
\label{eq:gbar_approx}
\end{equation} as derived in Appendix~\ref{app:score}.}

\begin{figure}[h]
\centering
\begin{subfigure}[t]{0.48\linewidth}
    \centering
    \includegraphics[width=\linewidth]{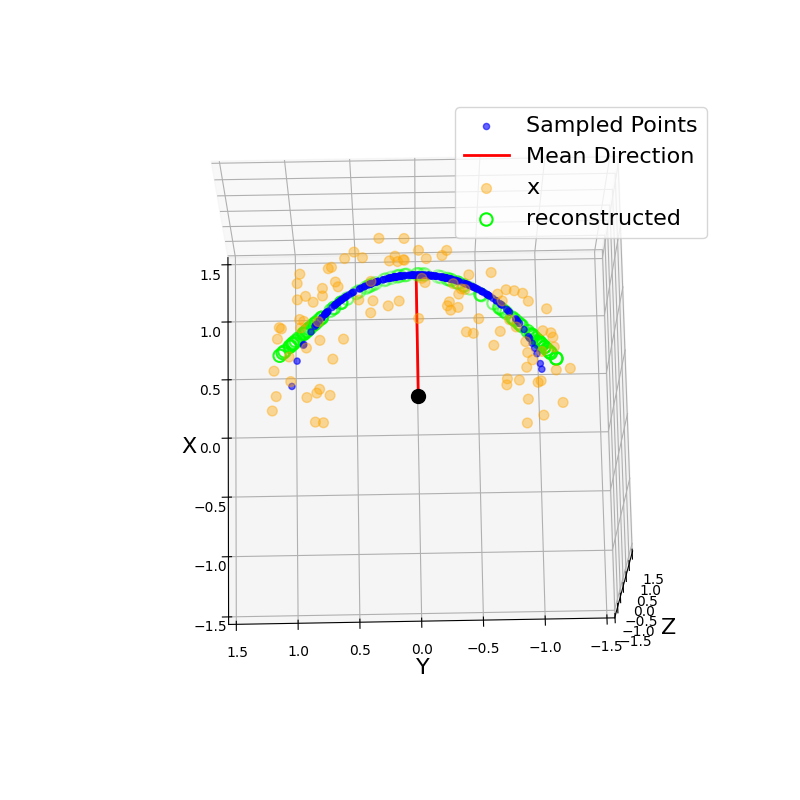}
    \vspace{-1.5cm}
    \caption{DAE restoration. MSE = 0.032, max error = 0.147.}
    \label{fig:G-bar-x-star_a}
\end{subfigure}
\hfill
\begin{subfigure}[t]{0.48\linewidth}
    \centering
    \includegraphics[width=\linewidth]{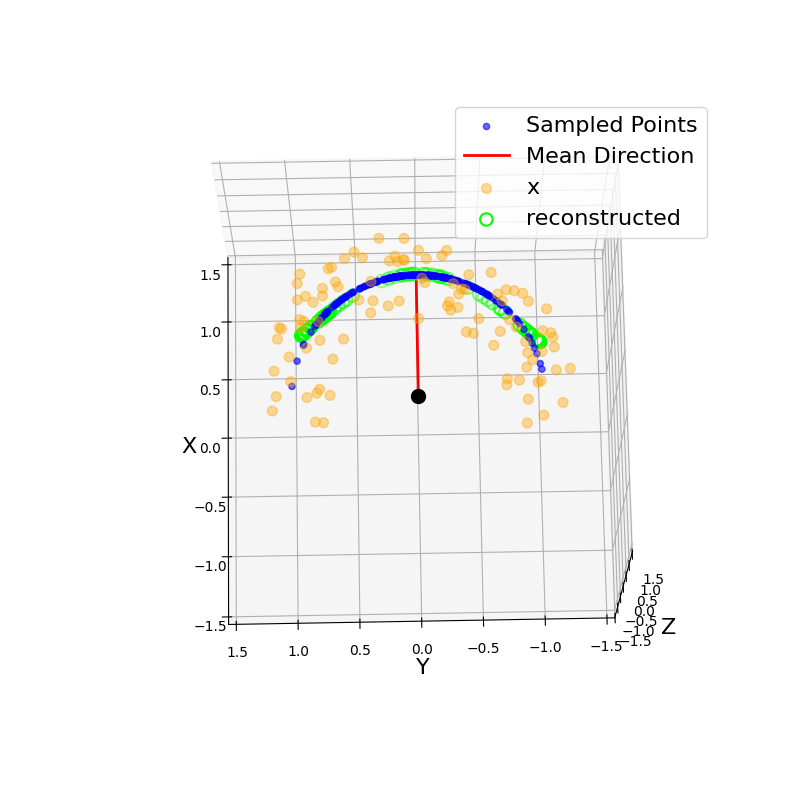}
    \vspace{-1.5cm}
    \caption{MPPM restoration. MSE = 0.026, max error = 0.060.}
    \label{fig:G-bar-x-star_b}
\end{subfigure}
\vspace{-0.4cm}
\caption{Comparison between denoising autoencoder (DAE) and our proposed MPPM, this example uses the same setup as in Fig.~\ref{fig:circle-illustration}. The error was computed as the deviation from the unit circle in 2D. In regions of the circle with lower probability density, the DAE is more prone to error than the proposed MPPM method.}
\label{fig:G-bar-x-star}
\end{figure}

 Algorithm~\ref{alg:mppm} outlines the training procedure using the clean dataset $\mathcal{X}^{\text{clean}}$ and the reconstruction of a noisy point $x$ in the ambient space. The algorithm is demonstrated for the simple case of a non-uniform distribution on the circle embedded in $\mathbb{R}^3$ in Figs.~\ref{fig:circle-illustration} and~\ref{fig:G-bar-x-star}, alongside a comparison with a vanilla denoising autoencoder (DAE). All the experimental and optimization details can be found in appendices~\ref{app:exp_setup} and~\ref{app:train_and_eval}.

\begin{algorithm}
\caption{MPPM}
\label{alg:mppm}
\begin{algorithmic}
\Function{Train}{$\mathcal{X}^{\text{clean}},\epsilon\sim \mathcal{N}(0,\sigma^2_d)$} 
\State $G, F, \di_\mi \gets \text{Train}\big(\mathcal{X}^{\text{clean}},\epsilon,\mathcal{L}(F,G,\di_\mi)\big)$ 
\EndFunction
\Function{Reconstruction}{${x},\mathcal{X}^{\text{clean}},\alpha,\beta$,num\_steps} 
\Comment {$0<\alpha,\beta,\alpha+\beta<1$}
\State {$x^1\gets {x}$}
 \For{$n \gets 1$ to num\_steps} 
 \State{$x^{n+1} \gets (1-\beta)x^n + \beta \sum_\alpha\bar{G}_\alpha(x^n) - \alpha \di_\mi(x^n)\hat{\nabla}_x\di_\mi(x^n)$} 
 \Comment{by \eqref{eq:x_follow_D_G_avg},\eqref{eq:gbar_approx}}
  \EndFor
   \State \Return {$x^{n+1}$}
\EndFunction
\end{algorithmic}
\end{algorithm}

\begin{figure}[ht]
\centering
\includegraphics[width=0.48\linewidth,trim=110 70 110 100,clip]{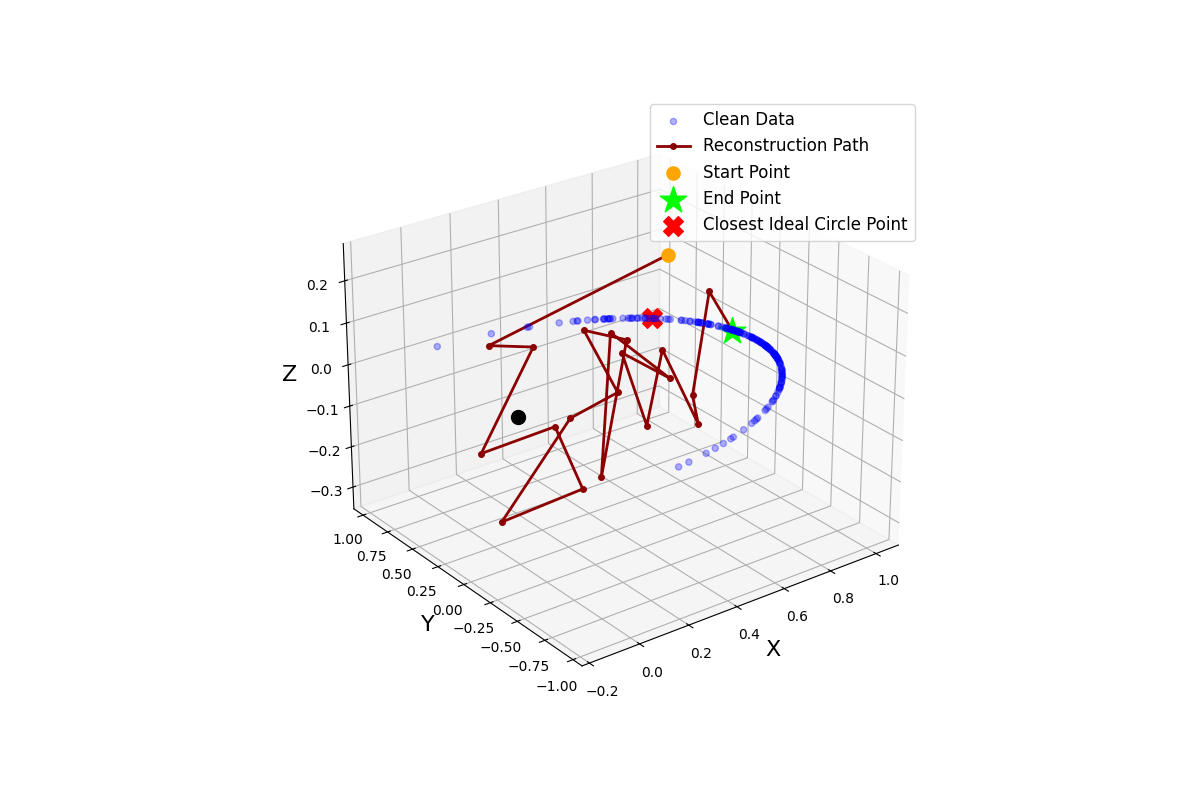}
\includegraphics[width=0.48\linewidth,trim=110 70 110 100,clip]{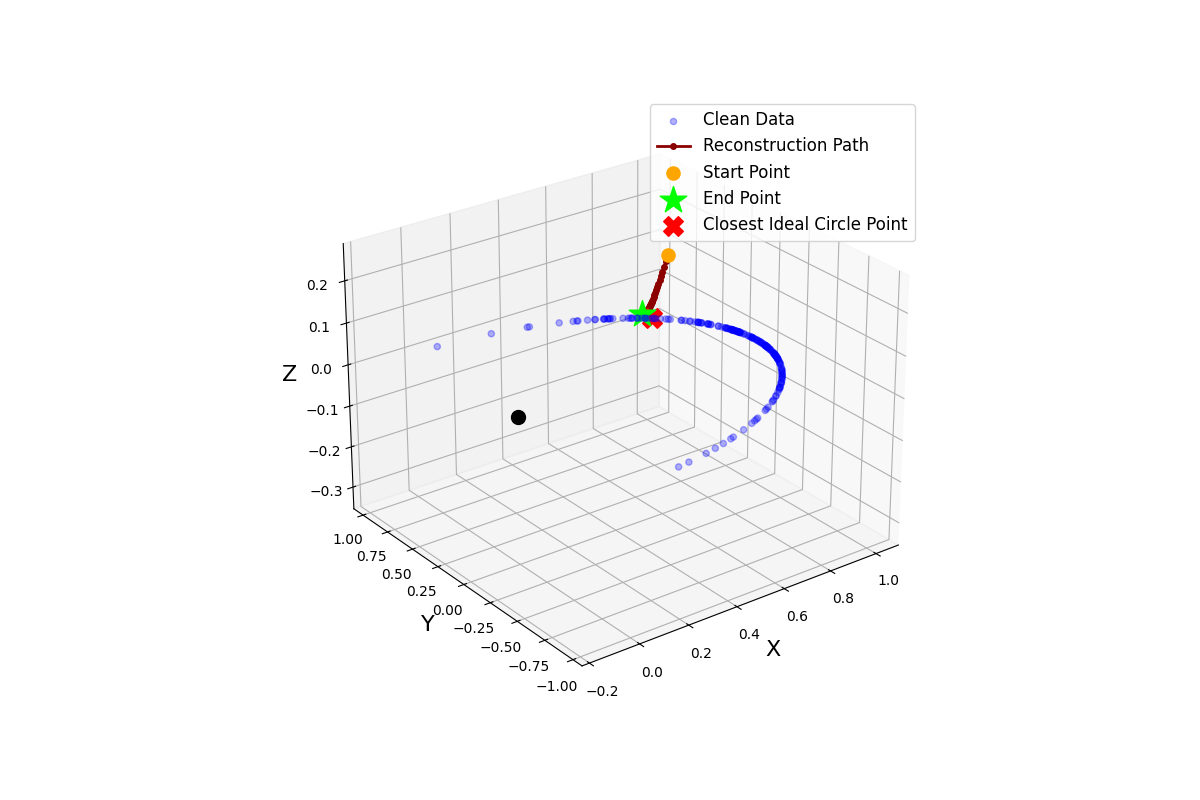}
\caption{Left: Diffusion model trajectory, Right: MPPM trajectory}
\label{fig:diff_vs_mppm}
\end{figure}
\begin{figure}[ht]
\centering
\includegraphics[width=0.48\linewidth,trim=110 0 110 10,clip]{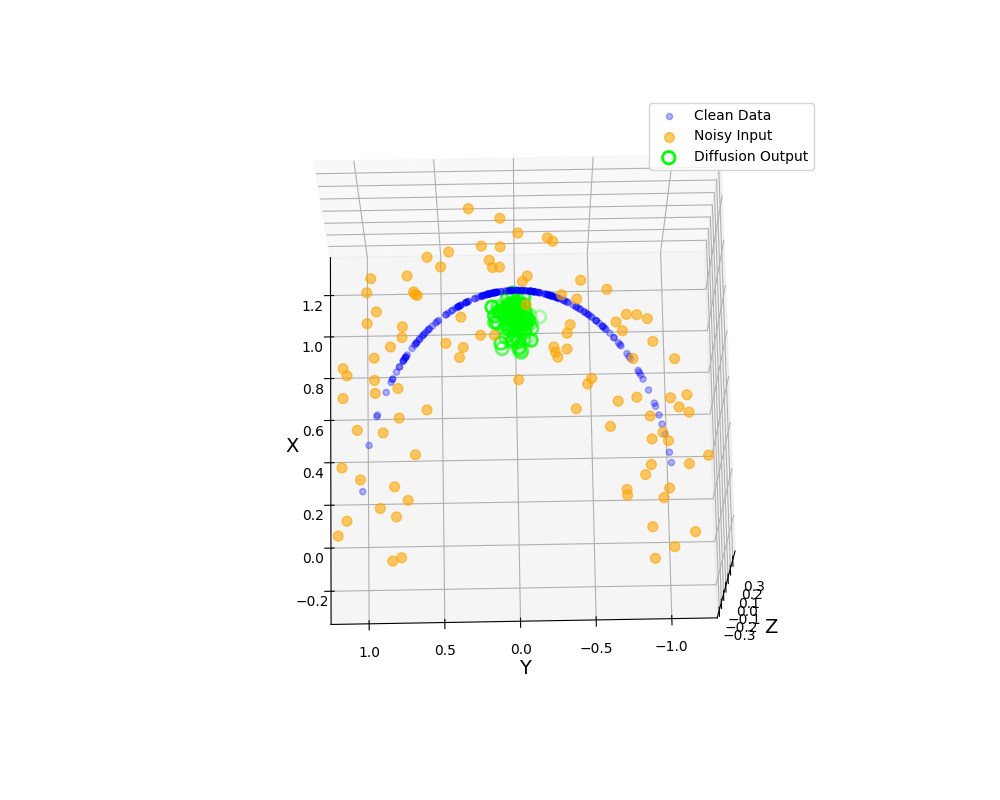}
\includegraphics[width=0.48\linewidth,trim=110 0 110 10,clip]{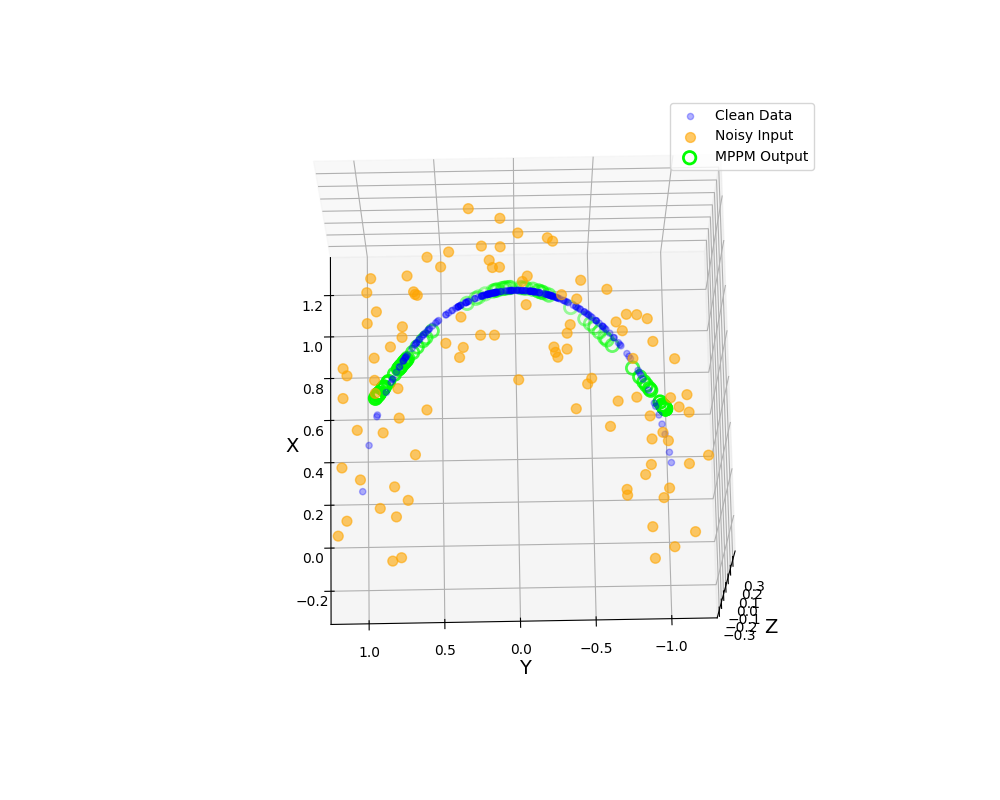}
\vspace{-1.5cm}
\caption{Left: Diffusion restoration, MSE=0.138, Right: MPPM restoration, MSE=0.026}
\label{fig:diff_vs_mppm_all_points}
\end{figure}

Figures~\ref{fig:diff_vs_mppm} and~\ref{fig:diff_vs_mppm_all_points} illustrate the advantages of the proposed MPPM method compared to the diffusion model. We used 1000 diffusion steps during both training and inference. Because the data points are not uniformly distributed, most of the reconstructed samples of the diffusion model are concentrated in the dense region (the upper half-circle), as shown in the left panel of Fig.~\ref{fig:diff_vs_mppm_all_points}. In contrast, the proposed MPPM method effectively handles this non-uniformity through our formulation, resulting in a significantly smaller reconstruction error. For completeness, the pseudocode for DDPM (Denoising Diffusion Probabilistic Model) training and inference is provided in Algorithms~\ref{alg:diff_train} and~\ref{alg:diff_inference} in Appendix~\ref{app:ddpm}.

\subsection{Latent MPPM (LMPPM)}
The key distinction between pixel space and latent space is that, in the latter, we do not assume that encoded clean and meaningful images lie on a lower-dimensional manifold. Instead, we regard the set of encoded clean and meaningful images as a point cloud that occupies the full latent space. We model this point cloud as samples from a probability distribution $P(z)$. Let $\mathcal{X}^{\text{clean}}$ denote the set of clean and meaningful images, and let
$S:=F\left(\mathcal{X}^{\text{clean}}\right)$
denote their corresponding latent representations. In this setting, $S$ plays the role previously assigned to the manifold $\mathcal{M}$. Specifically, the distance function $\di_S:\mathbb{R}^d\to \mathbb{R}$ is defined in the latent space with respect to the set of points $S$. Let $x\in\mathbb{R}^D$ denote an image and $z=F(x)\in\mathbb{R}^{d}$ its latent representation. Let $x^{\text{clean}}$ denote a clean image, and let $$x=x^{\text{clean}}+\epsilon,~~~\epsilon\sim \mathcal{N}(0,\sigma_{\text{noise}}I).$$
be the corresponding latent representations.
$$
z=F(x),~~~~~~z^{\text{clean}}=F(x^{\text{clean}}).
$$
The objective is to move the noisy latent representation $z$ toward a shifted representation $z^{\text{shift}}$ such that decoding it with $G$ produces a clean image. The loss consists of distance loss and reconstruction loss  for points inside and outside $S$, Specifically,
\begin{equation}
\mathcal{L} = \mathcal{L}_{\text{dist}} + \mathcal{L}_{\text{reco}},
\label{eq:total_loss}
\end{equation}
where
\begin{equation}
\mathcal{L}_{\text{dist}}=\sum_{z\in S}\left(\di_s(z^{\text{clean}})\right)^2+\sum_{z\notin S}\left(\di_s(z)-{\rm dist}_s(z,z^{\text{clean}})\right)^2,
\label{eq:loss_distance}
\end{equation}
and
\begin{equation}
\mathcal{L}_{\text{reco}}=\sum_{z\in S}\mathcal{L}_R(x^{\text{clean}},G(z^{\text{clean}}))+\sum_{z\notin S}\mathcal{L}_R(x^{\text{clean}},G(z^{\text{shift}})).
\label{eq:reco}
\end{equation}
The first term in~\eqref{eq:loss_distance} enforces zero distance for clean latent representations. The second term defines the distance of a noisy latent representation from $S$ using its distance to the corresponding clean representation, measured by the metric ${\rm dist_s(\cdot,\cdot)}$. The first term in~\eqref{eq:reco} encourages the decoder to reconstruct the clean image from its clean latent representation, and the second term encourages the shifted latent representation to reconstruct the corresponding clean image.

We define the distance between a noisy latent representation and its clean counterpart as
$$
{\rm dist}_s(z,z^{\text{clean}}):=\frac{1}{\sqrt{d}}\|z-z^{\text{clean}}\|,
$$
where normalization by the latent-space dimension $d$ reduces the dependence of the distance magnitude on dimensionality and improves numerical stability. Following the Tweedie formulation in latent space, we define the shifted latent representation as
$$
z^{\text{shift}}=(1-\beta)z+\beta \bar{z}-\alpha\di_S(z)\nabla_z\di_S(z).
$$
Here, $\bar{z}$ is derived from the score function of $P_{\text{ker}}$~\eqref{eq:pker},
\begin{equation}
s_{\text{ker}}=\nabla_z\log P_{\text{ker}}(z) = -\frac{1}{\sigma_{\text{ker}}^2}(z-\bar{z}),
\end{equation}
where
\begin{equation}
\bar{z}=\frac{\sum_{\alpha\in S} z_\alpha\exp\left({-\frac{|z-z_\alpha|^2}{2\sigma_{\text{ker}}^2}}\right)}
{\sum_{\alpha\in S} \exp\left({-\frac{|z-z_\alpha|^2}{2\sigma_{\text{ker}}^2}}\right)}.
\end{equation}
We approximate $\bar{z}$ as a soft local average of the nearest clean training latent representations, weighted by a Gaussian kernel. Following the approximation in~\eqref{eq:gbar_approx}, adapted to the latent space, we additionally normalize the squared distance by the latent-space dimension $d$ for numerical stability. We restrict the computation to the $k$ nearest neighbors. Let $\mathcal{N}_k(z)$ denote the index set of the $k$ nearest clean training samples to $z$:
$$
\mathcal{N}_k(z):=\underset{j}{\text{$k$-argmin}}
\|z-z_j^{\text{clean}}\|.
$$
Then,
\begin{equation}
\bar{z} =
\frac{\displaystyle\sum_{j\in \mathcal{N}_k(z)}
z_j^{\text{clean}}
\exp\left(-\big\lVert z-z_j^{\text{clean}}\big\rVert_2^2/
(d\cdot2\sigma_{\text{ker}}^2)\right)}
{\displaystyle\sum_{j\in \mathcal{N}_k(z)}
\exp\left(-\big\lVert z-z_j^{\text{clean}}\big\rVert_2^2/
(d\cdot2\sigma_{\text{ker}}^2)\right)}.
\label{eq:zbar_nn}
\end{equation}
We distinguish between geometric data, such as MNIST digits, and natural images, such as faces. For geometric data, the reconstruction loss is simply the squared $L_2$ distance between a reconstructed sample $p$ and its target $t$. For natural images, we additionally impose local geometric consistency in the latent space. In particular, for face images, we first assume that the faces are aligned. The proposed loss then encourages the latent space to be locally Euclidean, so that decoding the shifted representation $G(z^{\text{shift}})$ produces a valid face. For natural images, the reconstruction loss combines several complementary terms:
\begin{equation}
\mathcal{L}_R(p,t)=\|p-t\|^2+\lambda_{\text{per}}\mathcal{L}_{\text{per}}(p,t)+\lambda_{\text{ssim}}[1-\text{SSIM}(p,t)].
\label{eq:loss_rec}
\end{equation}
The first term is the pixel-wise $L_2$ reconstruction loss, the second is a perceptual loss, and the third encourages structural similarity through the SSIM measure \citep{wang2004image}  Unlike pixel-wise losses, perceptual loss compares images in the feature space of a pretrained deep network. Specifically, both images are passed through the network, and the discrepancy between their feature activations at selected layers is minimized. These features capture higher-level properties such as texture, structure, and semantic content, making perceptual loss more consistent with perceptual image similarity than a purely pixel-wise loss. The perceptual loss is defined as
$$
\mathcal{L}_{\text{per}}(p, t) \;=\; \frac{1}{|\mathcal{A}|} \sum_{\ell \in \mathcal{A}} \frac{1}{N_\ell} \sum_{u=1}^{N_\ell} \Big(\phi_\ell(p)_u - \phi_\ell(t)_u\Big)^2,
$$
where $\phi_\ell(\cdot)$ denotes the activation of VGG16 at layer $\ell$, and $$\mathcal{A}=\{3,8,15\}.$$
Following~\cite{wang2026}, and for improved numerical stability, we replace the gradient of the distance function by a direction field
$$V:\mathbb{R}^d\to\mathbb{R}^d$$
that points from the noisy latent representation $z$ toward its corresponding clean representation $z^{\text{clean}}$. For image data, we approximate the gradient by
$$
\nabla \di_S(z)\approx {{V}}(z)=\frac{z-z^{\text{clean}}}{\frac{1}{\sqrt{d}}\|z-z^{\text{clean}}\|+\varepsilon}.
$$
For face reconstruction, we further introduce an identity-preservation loss. This term encourages the reconstructed image to preserve the identity of the target subject independently of pixel-level fidelity. Specifically, the reconstructed and target images are passed through a pretrained face-recognition network, producing embedding vectors $e_p$ and $e_t$ respectively. The identity loss is defined as 
$$
\mathcal{L}_{\text{id}}(p,t)=1-\frac{e_p \cdot e_t}{\|e_p\|\|e_t\|}.
$$
Finally, we jointly train the four networks: the encoder $F$, decoder $G$, distance function $\di_S$, and direction field $V$ using the following objective:
\begin{equation}
\begin{aligned}
\mathcal{L}(F,G,\di_S,V)
    &= \mathcal{L}_{\text{dist}}+\mathcal{L}_V+\mathcal{L}_{\text{reco}},\\
\mathcal{L}_{\text{dist}}
    &= \lambda_1\sum_{z\notin S}
    \left(\di_S(z) - \frac{1}{\sqrt{d}}\|z-z^{\text{clean}}\|\right)^2
    +\lambda_2\sum_{z\in S}|\di_S(z)|^2,\\
\mathcal{L}_V
    &= \lambda_3\sum_{z\notin S}
    \left(
    V(z)-\frac{z-z^{\text{clean}}}
    {\frac{1}{\sqrt{d}}\|z-z^{\text{clean}}\|+\varepsilon}
    \right)^2,\\
\mathcal{L}_{\text{reco}}
    &= \lambda_4\sum_{z\in S}
    \mathcal{L}_R\left(x^{\text{clean}},G(z^{\text{clean}})\right)
    +\lambda_5\sum_{z\notin S}
    \mathcal{L}_R\left(x^{\text{clean}},G(z^{\text{shift}})\right)\\
    &\quad+\lambda_6\sum_{z\notin S}
    \mathcal{L}_{\text{id}}\left(x^{\text{clean}},G(z^{\text{shift}})\right).
\end{aligned}
\label{eq:main_loss_latent}
\end{equation}
where
\begin{equation}
    z^{\text{shift}}=(1-\beta)z+\beta \bar{z}-\alpha\di_S(z)V(z),
    \label{eq:z_shift_uv}
\end{equation}
the reconstruction loss is given by~\eqref{eq:loss_rec} and $\bar{z}$ is defined in~\eqref{eq:zbar_nn}. 

During training, we uniformly sample different values of $\alpha$ and $\beta$ in order to expose the model to a variety of blending conditions and improve its robustness. During inference, specific values of $\alpha$ and $\beta$ are selected according to the degradation level of the input image. To determine the number of iterations required during inference, we additionally estimate the probability that the reconstructed image $G(z)$ belongs to the clean-image distribution. The iterative procedure is terminated when this probability reaches a plateau within a prescribed tolerance. For a latent representation $z$, let $z_1,\ldots,z_k$ denote its $k$ nearest training latent representations, where proximity is measured using the squared Euclidean distance in the latent tensor. We estimate the probability as the average of the corresponding local density estimates: 
\begin{equation}
P(z) = \frac{1}{k}\sum_{j=1}^{k} p_j(z).
\label{eq:prob_stop}
\end{equation}
For each neighbor $z_j$, we draw $N$ Monte Carlo samples
$$
z_j^{(m)} \sim \mathcal{N}(z_j, \sigma_{\text{ker}}^2 I), \qquad m=1,\ldots,N,
$$
decode them using $G$, and estimate the local Gaussian-kernel density in pixel space:
$$
p_j(z) = \mathbb{E}_{\tilde{z} \sim \mathcal{N}(z_j,\sigma_{\text{ker}}^2 I)}
\left[
\exp\left(
-\frac{\lVert G(\tilde{z})-G(z)\rVert^2/D}
{2\sigma_{\mathrm{pix}}^2}
\right)
\right].
$$
In practice, this expectation is approximated using the $N$ Monte Carlo samples. Here, $D=C\times H\times W$ denotes the image dimension, $\sigma_{\text{ker}}$ controls the extent of the perturbation around each training latent representation, and $\sigma_{\text{pix}}$ controls the bandwidth of the pixel-space similarity kernel. The complete procedure is summarized in Algorithm~\ref{alg:lmppm}.

\begin{algorithm} 
\caption{LMPPM}
\label{alg:lmppm}
\begin{algorithmic}
\Function{Train}{$\mathcal{X}^{\text{clean}}, \alpha,\beta\sim U(\cdot,\cdot)$}
\State $G, F, \mathcal{\di_S}, V \gets \text{Train}\big(\mathcal{X}^{\text{clean}},\alpha,\beta,\mathcal{L}(F,G,\mathcal{\di_S},V)\big)$
\Comment{by~\eqref{eq:zbar_nn},\eqref{eq:loss_rec},\eqref{eq:main_loss_latent},\eqref{eq:z_shift_uv}}
\EndFunction
\Function{Reconstruction}
{${x},\mathcal{X}^{\text{clean}},\alpha,\beta,\tau$,max\_steps} 
\Comment {$0<\alpha,\beta,\alpha+\beta<1$}
\State {$z^1\gets F({x})$}
 \For{$n \gets 1$ to max\_steps} 
 \State{$z^{n+1} \gets (1-\beta)z^n + \beta \bar{z}^n- \alpha \mathcal{\di_S}(z^n)V(z^n)$}
 \Comment{by~\eqref{eq:z_shift_uv}}
 \If{$|P(z^{n+1})-P(z^{n})| < \tau$ }
 \Comment{by \eqref{eq:prob_stop}}
    \State \textbf{break}
\EndIf
\EndFor
   \State \Return {$G(z^{n+1})$}
\EndFunction
\end{algorithmic}
\end{algorithm}

\section{Experiments}
\label{sec:exper}
We evaluate our MPPM method on synthetic data and our LMPPM method on real-world image datasets, with particular emphasis on challenging image restoration settings. For the facial datasets, we additionally compare our method with DiffBIR~\cite{DiffBIR}, a state-of-the-art blind image restoration approach based on latent diffusion, providing a strong modern baseline for evaluating restoration quality. Across all experiments, we simultaneously train compact networks for $F$, $G$, $V$, $\di_\mi$, and $\di_S$, with fewer than 6M parameters in total.

Importantly, our models are trained exclusively using Gaussian noise degradation, while at inference time they are evaluated under a diverse range of degradation types that were not encountered during training. This setup directly tests the ability of the proposed manifold-based prior to generalize beyond the degradation distribution used during training. We compare our results with standard denoising autoencoders (DAE)~\citep{denoising_original_vincent2008extracting} and latent diffusion models (LDM)~\citep{rombach2022ldm}.

For synthetic experiments, we consider a one-dimensional manifold consisting of a half-circle in the $xy$-plane embedded in $\mathbb{R}^3$. The points in the manifold are sampled according to angular coordinates drawn from truncated normal distributions (see Figs.~\ref{fig:circle-illustration} and~\ref{fig:G-bar-x-star}). For real-world experiments, we consider MNIST~\citep{lecun1998mnist}, the SCUT-FBP5500 facial beauty dataset~\cite{liang2017SCUT}, and CelebA-HQ-256~\cite{celeba}. In MNIST, we evaluate restoration under three distinct degradation types: Gaussian noise, downsampling (super-resolution), and elastic deformation, each at two severity levels. For SCUT-FBP5500 and CelebA-HQ-256, we consider four types of degradation: Gaussian noise, random scribbles, small black patches, and partial masking, including masking of image regions and the eyes. For both facial datasets, we further compared our method with DiffBIR~\cite{DiffBIR}, enabling evaluation against a powerful diffusion-based restoration framework. Detailed architecture specifications and hyperparameters are provided in Appendices~\ref{app:exp_setup} and~\ref{app:train_and_eval}.

\begin{figure}[ht]
    \centering
    \includegraphics[width=0.30\linewidth]{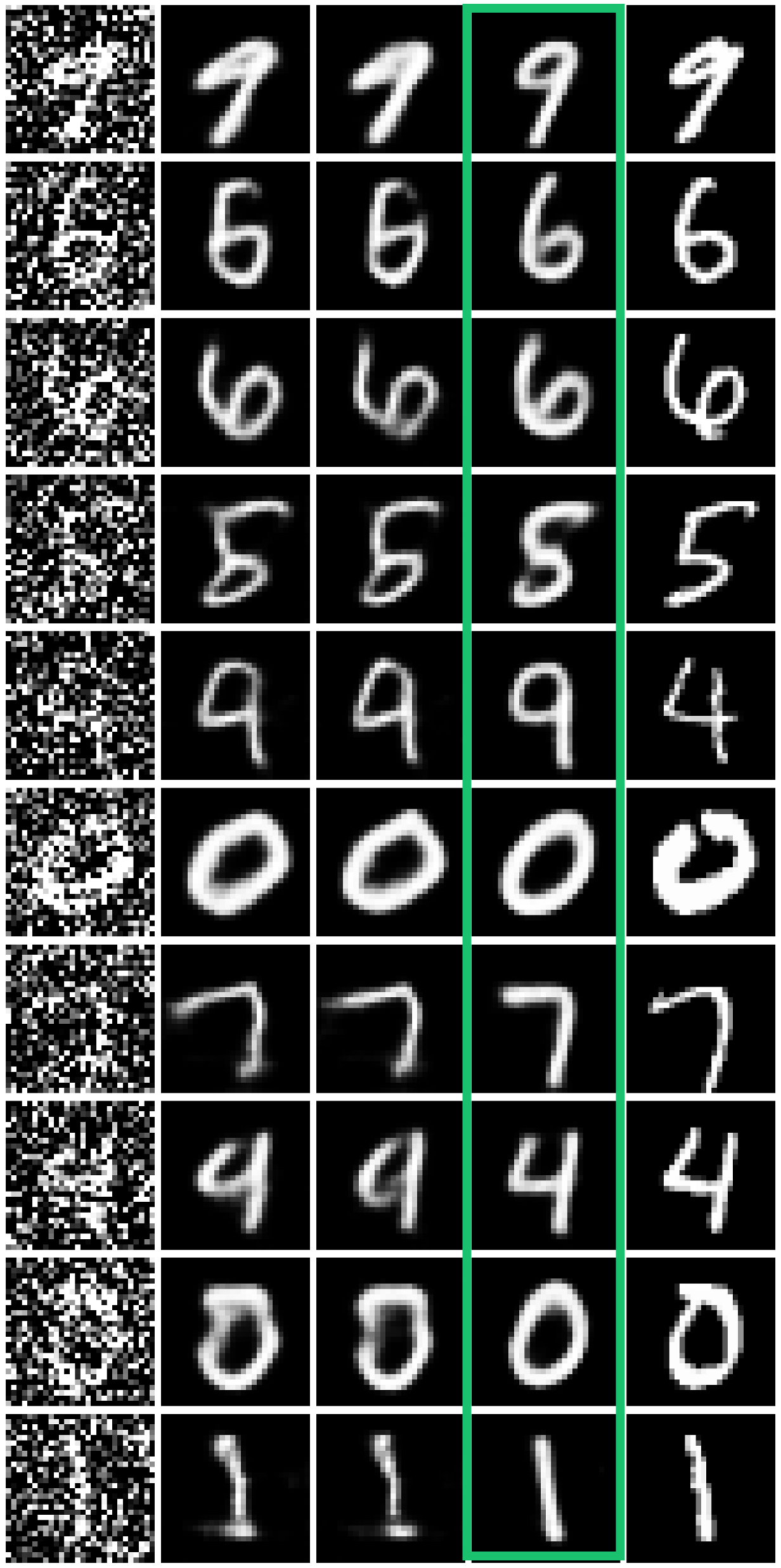}\hfill
    \includegraphics[width=0.30\linewidth]{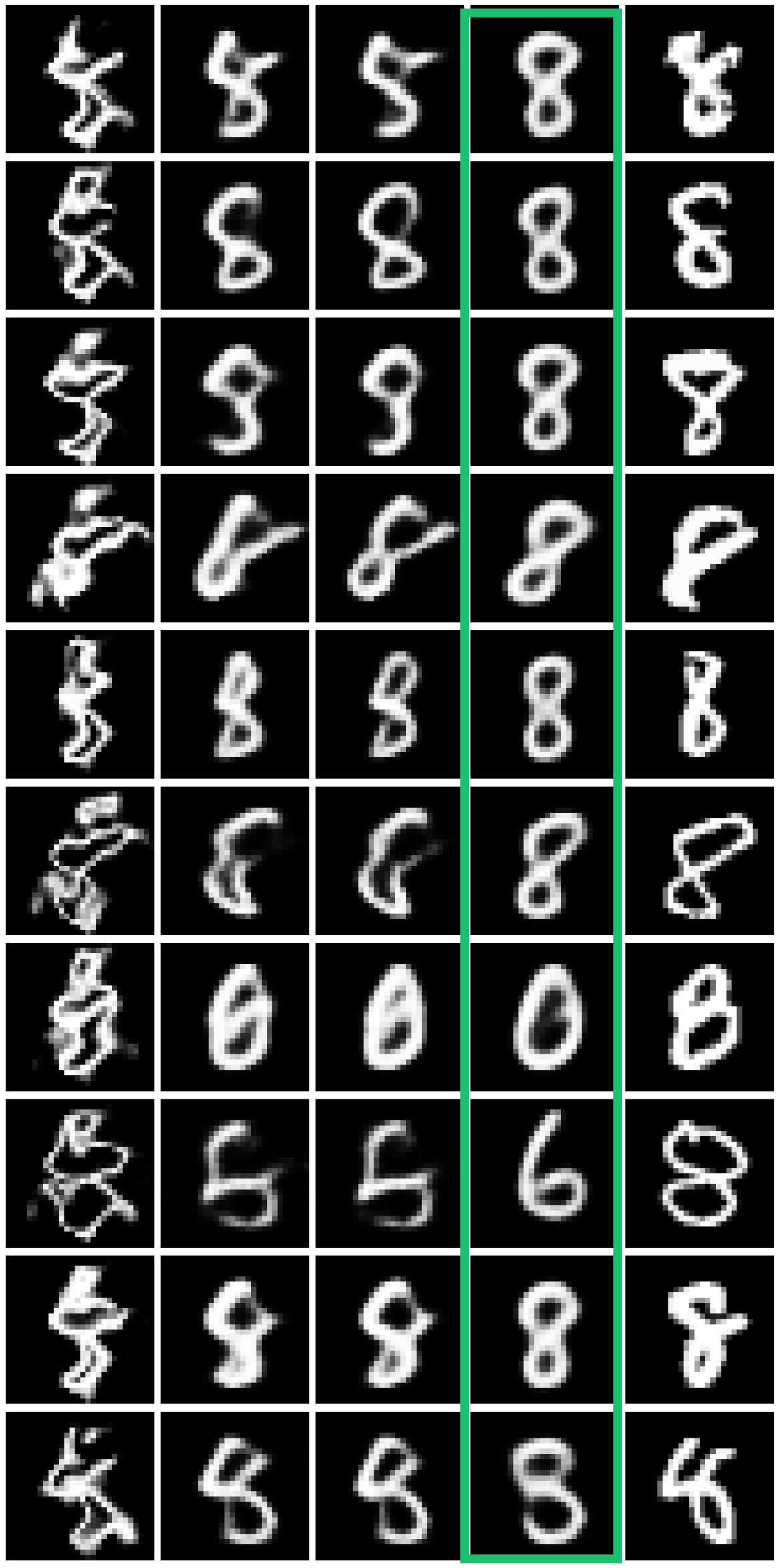}\hfill
    \includegraphics[width=0.30\linewidth]{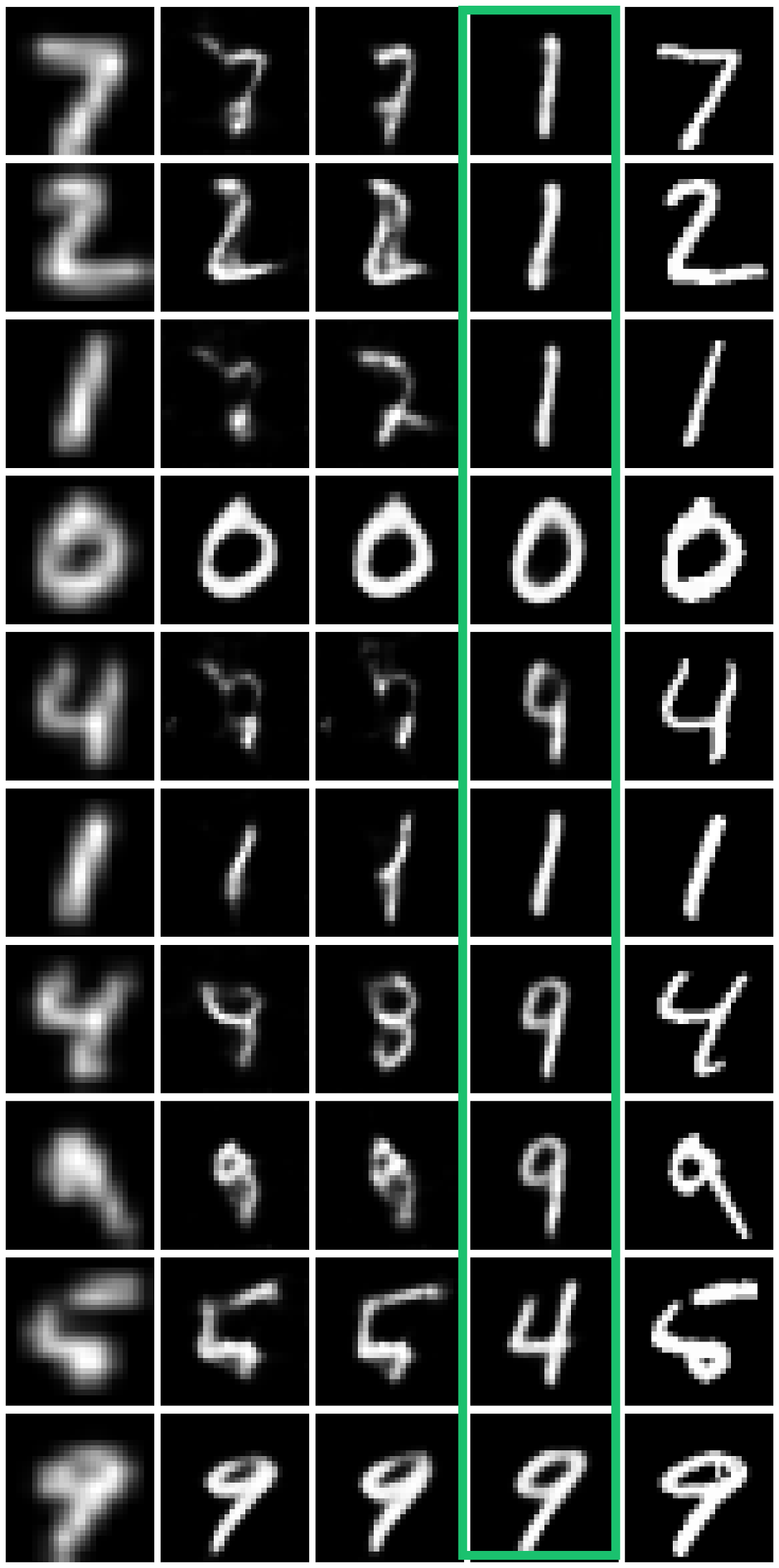}
   \caption{Left panel: noise $\sigma=0.7$; middle panel: elastic ($\alpha=0.34$, $\sigma=1.8$); 
right panel: downsampling factor $=0.35$. In all panels, from left to right: degraded, 
DAE, LDM, LMMPM (ours), and original.}
    \label{fig:mnist_defor}
\end{figure}
\begin{figure}
    \centering
    \includegraphics[width=0.9\linewidth]{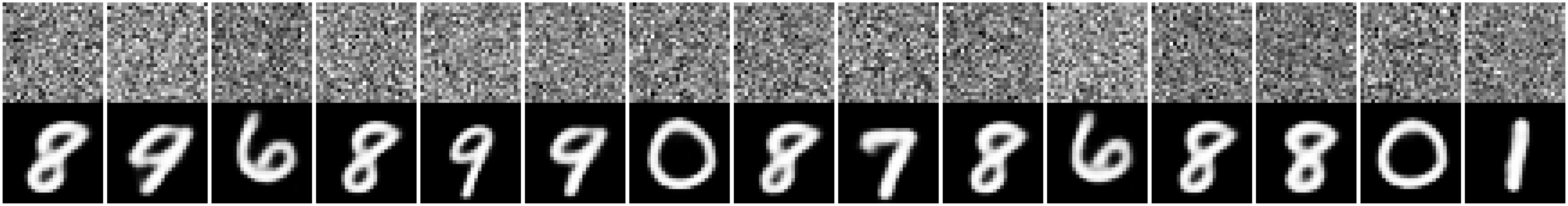}
    \includegraphics[width=0.92\linewidth]{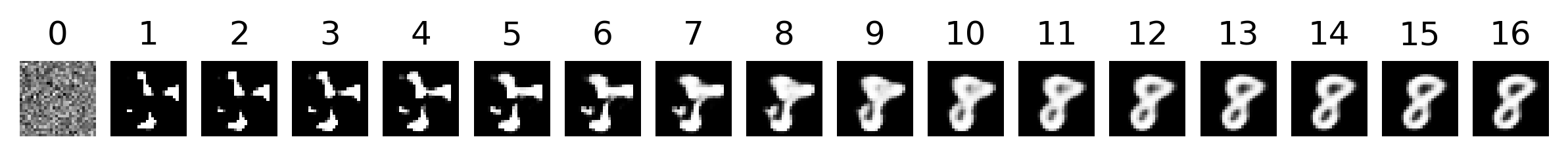}
    \vspace{-6pt}
    \caption{Top: Digit generation from pure  noise, with an FID of 19.53 computed over 2000 images. Bottom: Progression of digit generation over 16 steps.}
    \label{fig:noise2image}
\end{figure}

For the synthetic data, we implement MPPM using MLP architectures. For MNIST, we employ a CNN-based autoencoder for both DAE and our LMPPM method, while for SCUT-FBP5500 and CelebA-HQ-256 we adopt convolutional and U-Net-like architectures (see Appendix~\ref{app:exp_setup} for details). For LDM applied to MNIST, we integrate the corresponding DAE backbone (in place of the autoencoder) with a standard diffusion model, using 2000 diffusion steps. For LDM applied to face images, we use an SDEdit-style approach with Stable Diffusion v1.5: input images are encoded into the VAE latent space, partially noised according to a configurable strength parameter (rather than fully noised), and then denoised using a DDIM scheduler. The resulting latents are decoded back through the VAE and resized to the original image resolution.
\subsection{Results} 
\label{sec:results}
{\bf MNIST results:}
For the MNIST dataset, we set the latent space dimension to 18 and the additive noise to $\epsilon=0.4$. To calculate FID, we trained an MNIST classifier and computed an embedding distribution for each class. After reconstructing a degraded digit, we classified it and compared its embedding with the corresponding pre-computed class distribution. Table~\ref{tab:mnist_results} reports the mean SSIM and FID metrics. Our method consistently outperforms both DAE and LDM baselines across all degradation types in terms of FID scores. Notably, DAE occasionally achieved higher SSIM values, although its visual results were inferior.
We additionally performed an ablation study to assess the significance of the distance network. Ablation$^{\text{lmppm}}$ corresponds to setting $\alpha = 0$ in the reconstruction process, while Ablation$^{\text{dae}}$ uses the DAE network instead of our $(F, G)$ network, also with $\alpha = 0$. As can be seen, when using the proposed network (trained with $\mathcal{D}$), the results improve compared to the DAE variant, but still remain below the performance of the full reconstruction setting ($\alpha > 0$).
 
Fig.~\ref{fig:mnist_defor} illustrates restoration examples for Gaussian noise, elastic deformation, and downsampling. Additional experiment included the generation of digits from a pure noise. We generated 200 images from random Gaussian noise and managed to obtain realistic digits (FID=19.5) as can be seen in Fig.~\ref{fig:noise2image}. 

\begin{figure}
\centering
 \includegraphics[width=0.9\linewidth]{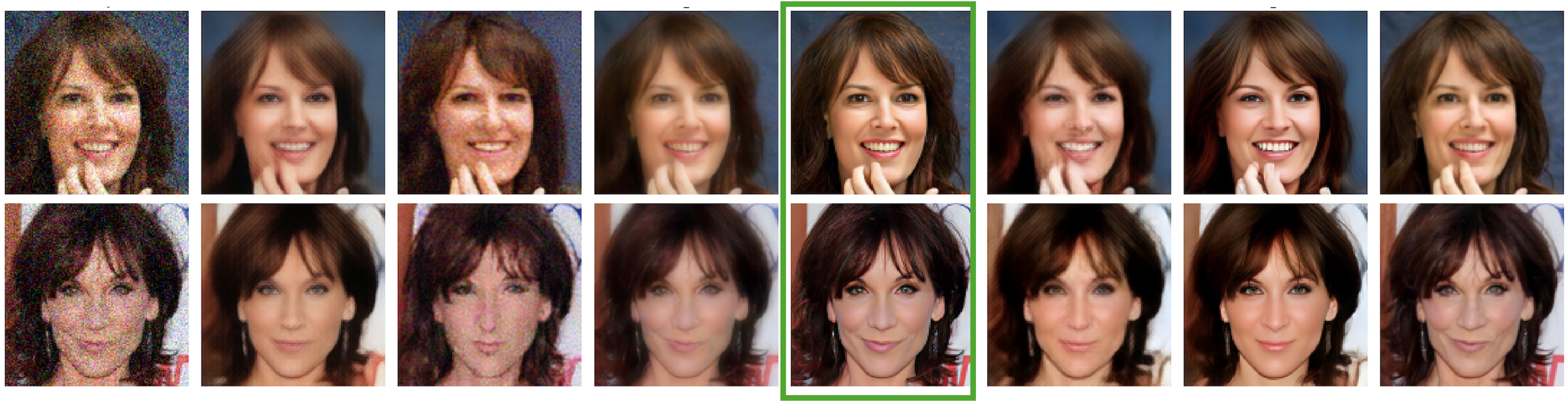}
    \caption{Noise $\sigma=0.1$ applied on CelebA-HQ-256. From left to right: degraded, DAE, LDM, DiffBir stage 1,DiffBir, LMPPM, LMMPM +DiffBir stage2, original.}
    \label{fig:noise_0.1}
\end{figure}

\begin{figure}
    \centering
    \includegraphics[width=0.9\linewidth]{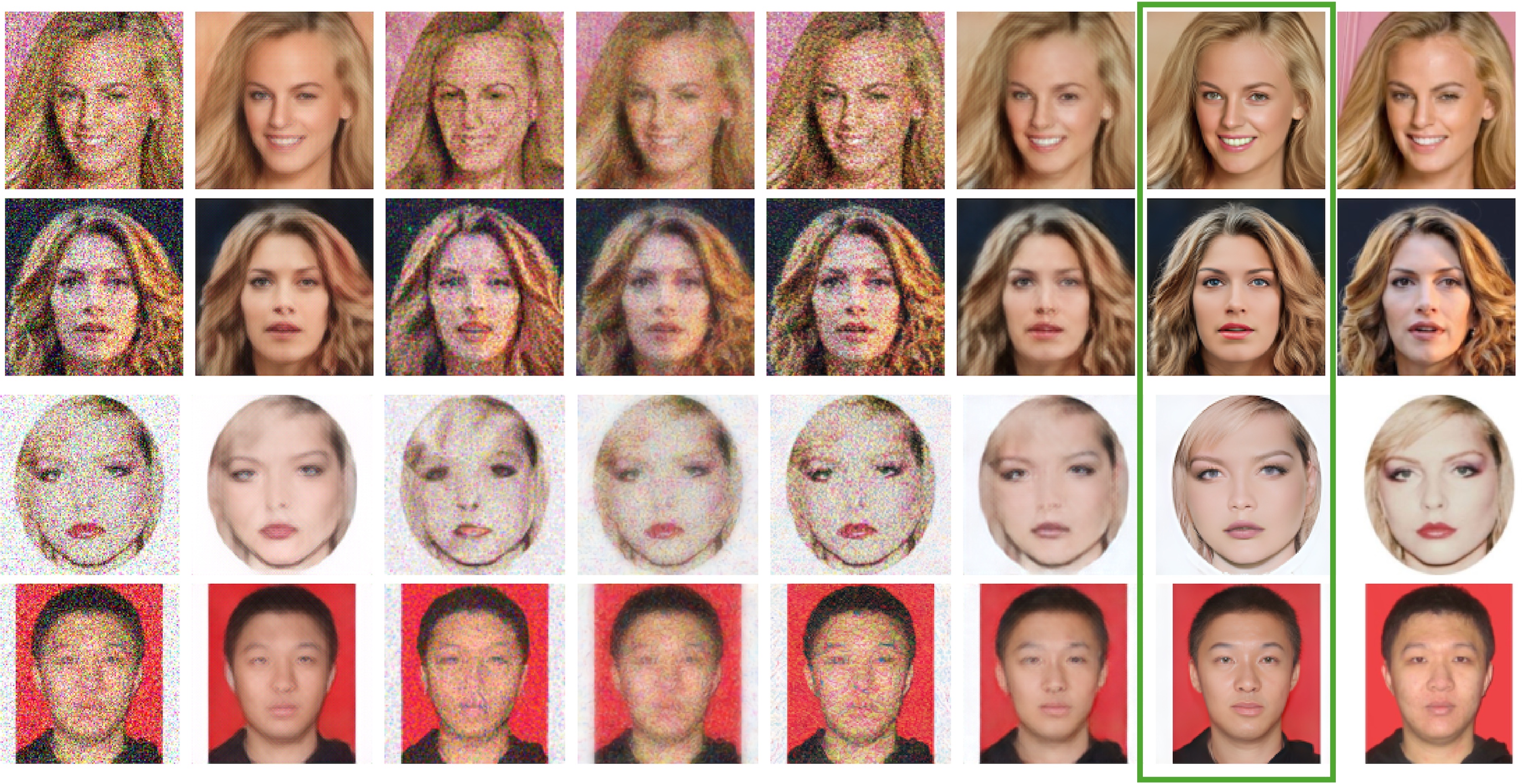}
    \caption{Noise $\sigma=0.2$. From left to right: degraded, DAE, LDM, DiffBir stage 1, DiffBir, LMPPM, LMMPM +DiffBir stage2, original.}
    \label{fig:noise_0.2}
\end{figure}

\begin{figure}
    \centering
   \includegraphics[width=0.9\linewidth]{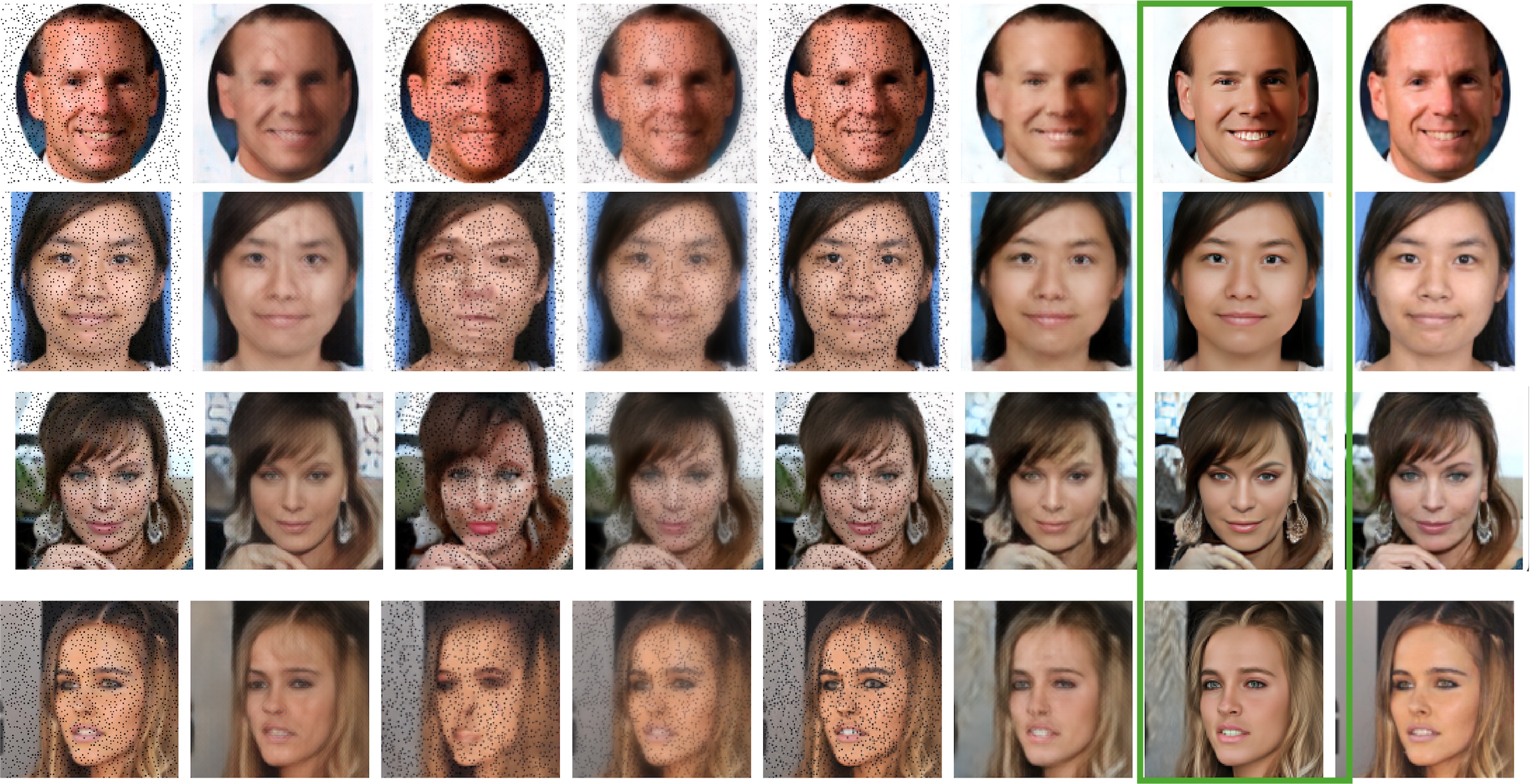}
   \caption{Missing pixels (0.08). From left to right: degraded, DAE, LDM, DiffBir stage 1, DiffBir, LMPPM, LMMPM +DiffBir stage2, original.}
    \label{fig:patches_0.08}
\end{figure}

\begin{figure}
    \centering
    \includegraphics[width=0.9\linewidth]{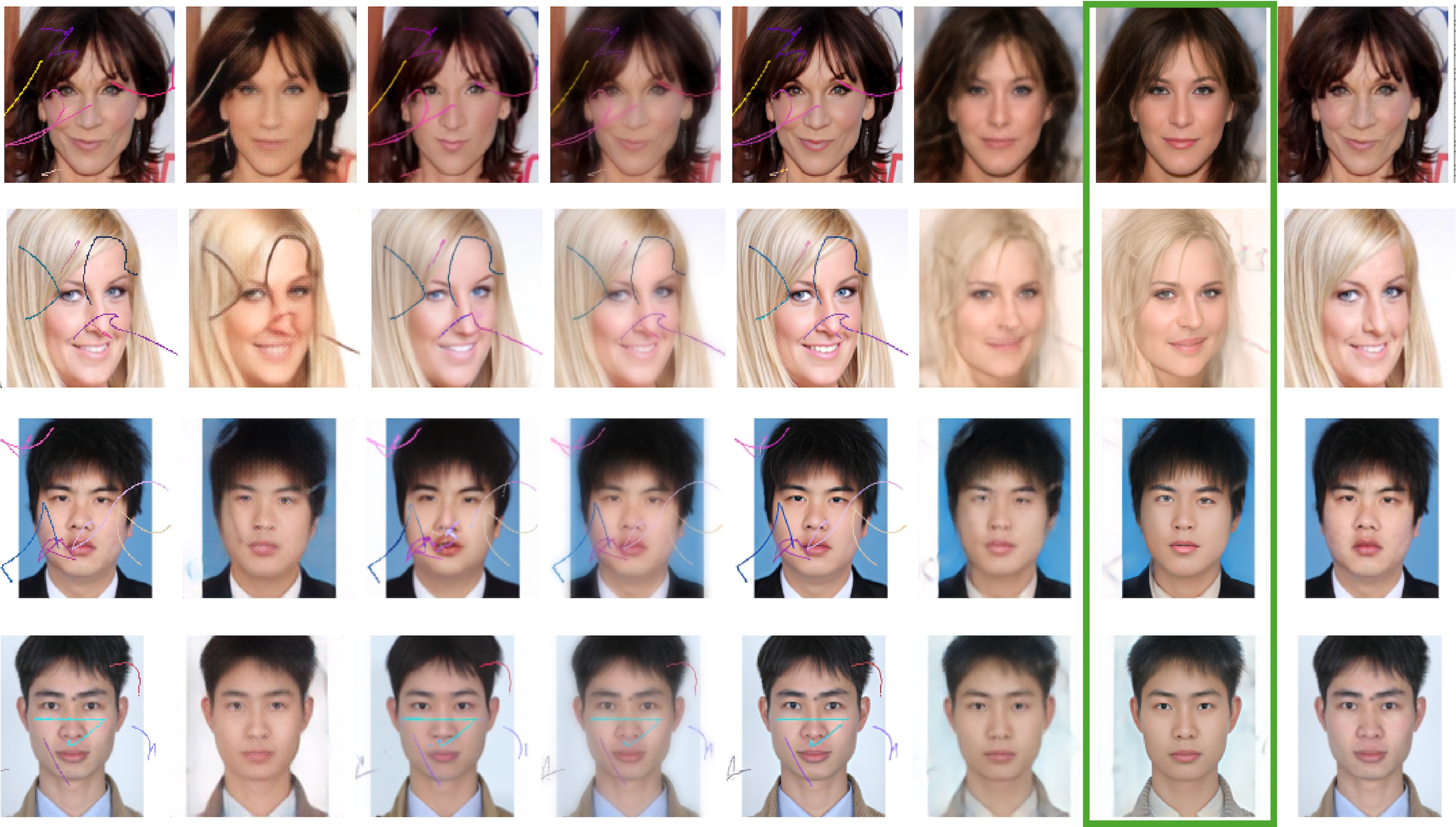}
    \caption{ 
Images with 5 scribbles. From left to right: degraded, DAE, LDM, DiffBir stage 1, DiffBir, LMPPM, LMMPM +DiffBir stage2, original.}
    \label{fig:scribble_5}
\end{figure}

\begin{figure}
    \centering
   \includegraphics[width=0.9\linewidth]{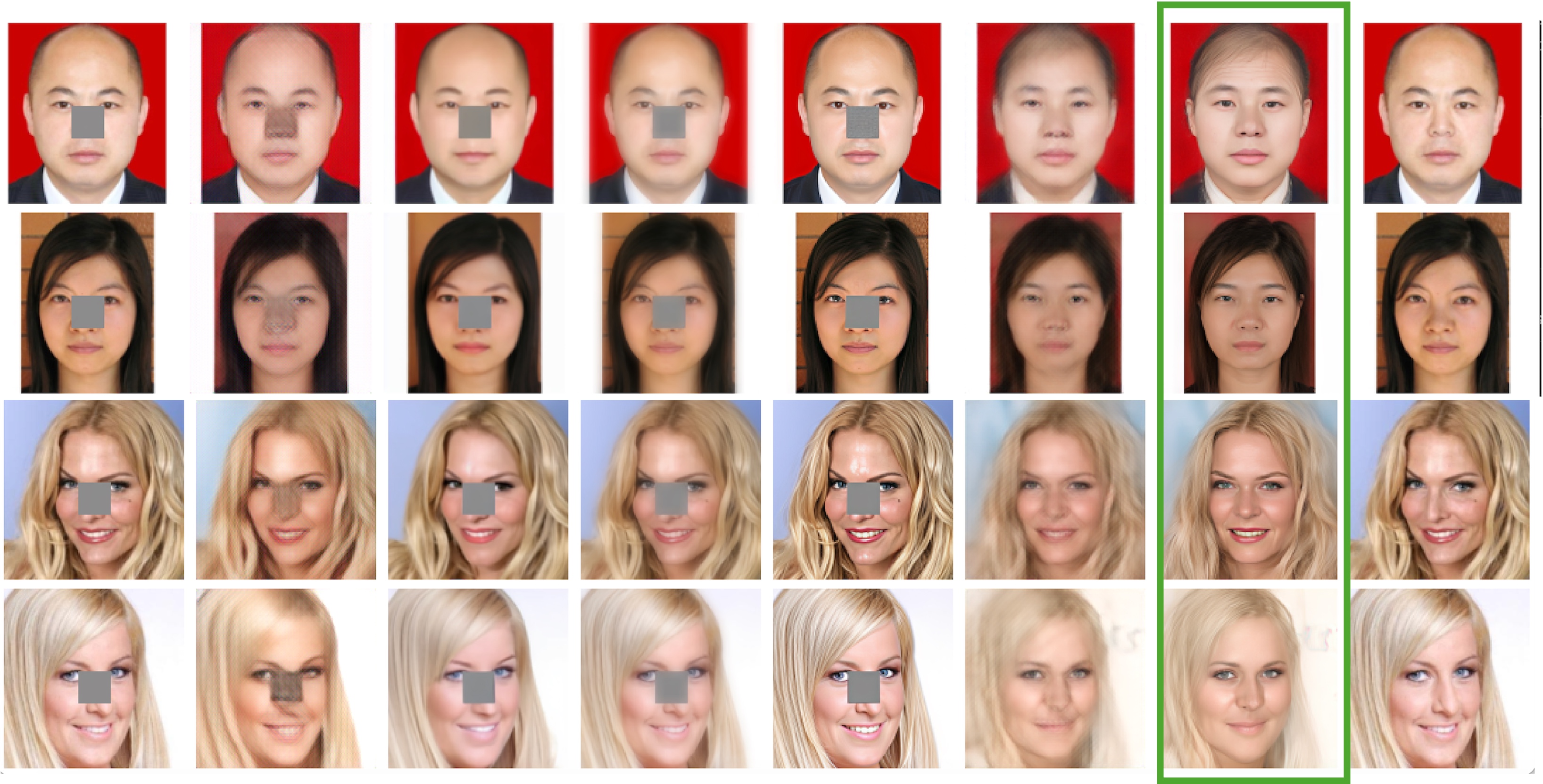}
   \caption{Nose Masking. From left to right: degraded, DAE, LDM, DiffBir stage 1, DiffBir, LMPPM, LMMPM +DiffBir stage2, original.}
    \label{fig:nose_18}
\end{figure}

\begin{figure}
    \centering
   \includegraphics[width=0.9\linewidth]{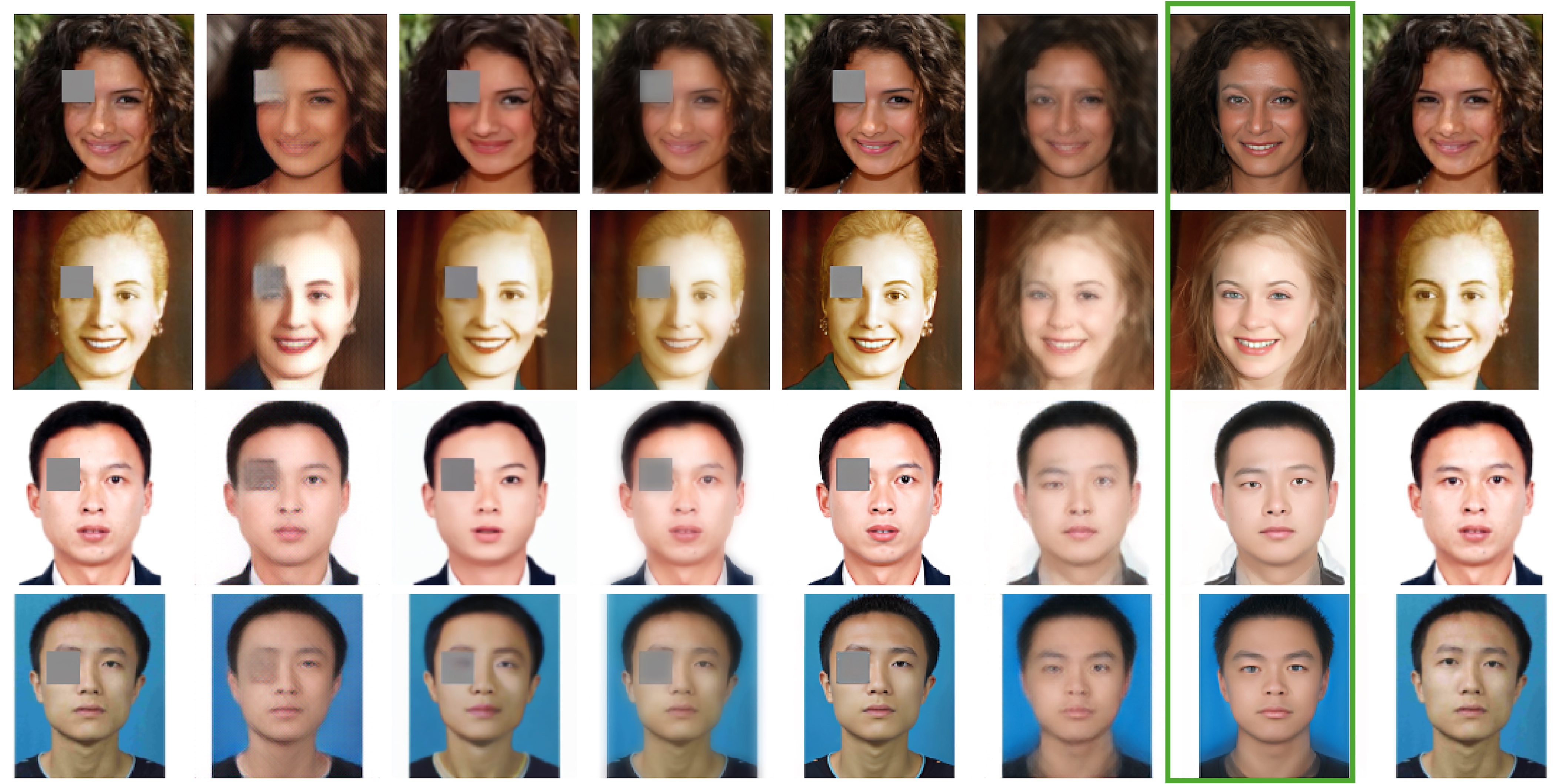}
   \caption{Eye Masking. From left to right: degraded, MPPM with no ID loss, MPPM, DiffBir stage 1, DiffBir, LMPPM, LMMPM +DiffBir stage2, original.}
    \label{fig:left_eye_18}
\end{figure}

{\bf Faces Results:}
Figs.~\ref{fig:noise_0.1}-\ref{fig:left_eye_18} present restoration results for facial images from both the SCUT-FBP5500 and CelebA-HQ-256 datasets under different types of degradation: randomly missing pixels, random scribbles, Gaussian noise, and missing facial parts (the nose and left eye). Note that, unlike inpainting methods, the missing mask was not provided to the algorithm. We evaluated the results using LPIPS and FID.

We further compared our method with DiffBIR~\cite{DiffBIR}, which addresses blind image restoration using two stages: degradation removal and information regeneration. The first stage removes the degradation and produces a high-fidelity but often over-smoothed intermediate result, while the second stage regenerates realistic textures and details. For completeness, we conducted three experiments: (i) DiffBIR using only its first stage, (ii) the full DiffBIR pipeline, and (iii) our LMPPM followed by DiffBIR's second stage.

Quantitative results are shown in Tables~\ref{tab:faces} 
At a low Gaussian noise level $(\sigma=0.1)$, both the visual results (Fig.~\ref{fig:noise_0.1}) and the quantitative results show that DAE and DiffBIR outperform LMPPM, as expected, since they are specifically designed for this type of degradation. However, at higher noise levels, the quantitative evaluation supports the visual findings, with our method consistently achieving lower FID scores across all degradation scenarios.

For these experiments, we set the latent dimension to 1024 and the additive noise level to $\epsilon=0.2$. As can be seen, LMPPM outperforms the first stage of DiffBIR both visually and quantitatively. The output of DiffBIR's second stage is realistic and perceptually high-quality. Notably, the full DiffBIR model (fourth column from the right) performs well in removing Gaussian noise (first row), even though the reconstructed image differs from the original image. The best performance is achieved by applying our LMPPM followed by DiffBIR's second stage (second column from the right), indicating that our blind degradation-removal module provides a strong foundation for high-quality restoration.

\begin{figure}
    \centering
   \includegraphics[width=0.45\linewidth]{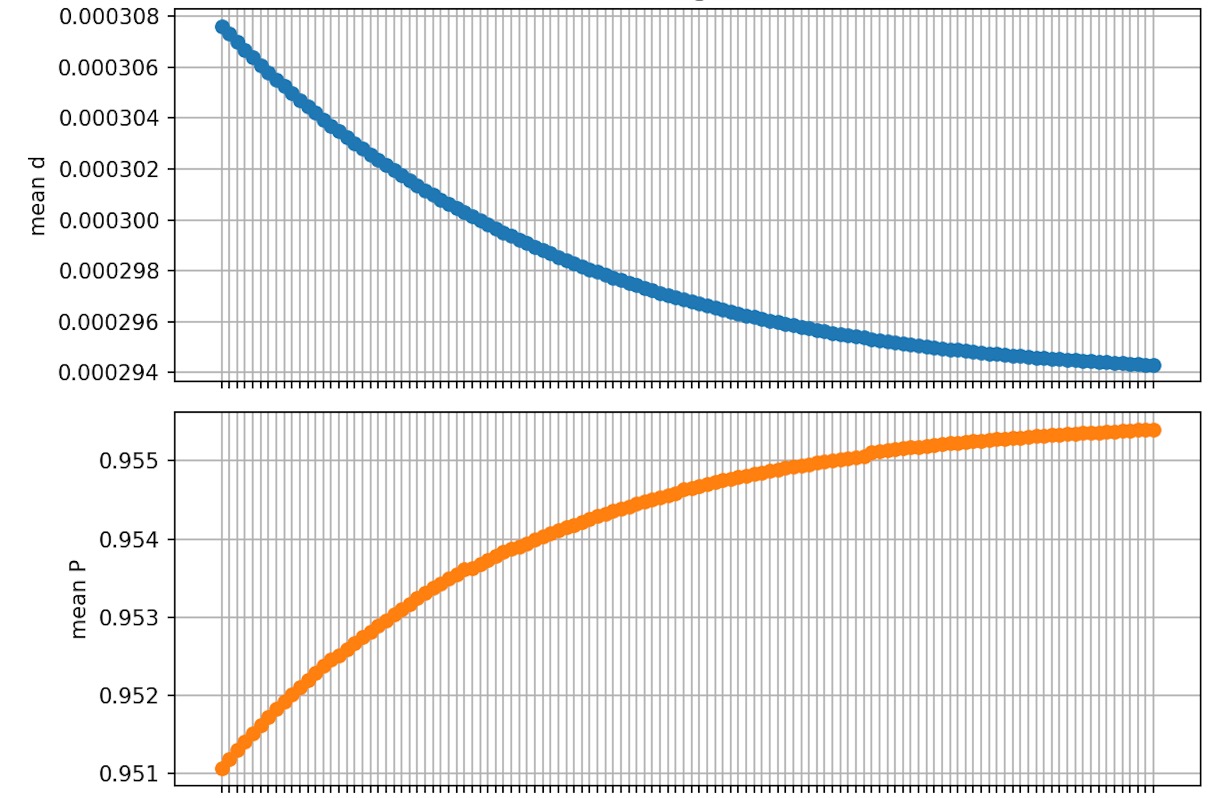}
   \caption{Inference convergence (scribble). Top: distance to manifold $\di_S$. Bottom: probability $P(z)$. Iterations terminate once probability plateaus.}
    \label{fig:convergence}
\end{figure}

Note that when parts of the face are masked, our method completes the missing regions using the $\bar{z}$ element, which incorporates information from the nearest neighbors in the latent space. The convergence of the algorithm is demonstrated in Fig.~\ref{fig:convergence}. The top panel shows the decay of the distance function $\di_S$, while the bottom panel shows the increase in $P(z)$. Both measures converge and reach a plateau.

\begin{table}
\begin{center}
\caption{Quantitative results on MNIST}
\label{tab:mnist_results}
\begin{tabular}{@{}l|ll|ll@{}}
\toprule
& \multicolumn{2}{c|}{Elastic 2.3}
& \multicolumn{2}{c}{Elastic 1.8}\\
& SSIM $\uparrow$& FID $\downarrow$& SSIM $\uparrow$& FID $\downarrow$\\
\toprule
DAE& \underline{0.66}& 69.36& \underline{0.59}& 134.60\\
LDM& 0.64& 66.52& 0.58& 124.05\\
LMPPM (ours)& 0.63& {\bf12.61}& \underline{0.59}& {\bf16.38}\\
\midrule
{Ablation$^{\text{lmppm}}$} & 0.63 & 12.83 & 0.59 & 16.27\\
{Ablation$^{\text{dae}}$}   & 0.17 & 522.39 & 0.15 & 527.25 \\
\midrule
\midrule
& \multicolumn{2}{c|}{Downsample 0.5}
& \multicolumn{2}{c}{Downsample 0.35}\\
& SSIM $\uparrow$& FID $\downarrow$& SSIM $\uparrow$& FID $\downarrow$\\
\toprule
DAE& \underline{0.79}& 31.66& \underline{0.54}& 133.66\\
LDM& 0.75& 31.61& 0.53& 128.80\\
LMPPM (ours)& 0.67& {\bf11.27}& 0.52& {\bf22.65}\\
\midrule
{Ablation$^{\text{lmppm}}$} & 0.67 & 11.34 & 0.52 & 22.89 \\
{Ablation$^{\text{dae}}$}   & 0.17 & 521.14 & 0.13 & 504.08 \\
\bottomrule
\end{tabular}
\end{center}
\end{table}

\begin{table}
\begin{center}
 \caption{Quantitative results on SCUT-FBP5500}
 \label{tab:faces}
\begin{tabular}{@{}l|ll|ll|ll@{}}
\toprule
& \multicolumn{2}{c|}{Noise 0.1}
& \multicolumn{2}{c|}{Noise 0.2}
& \multicolumn{2}{c}{Miss pixels 0.08}\\
& LPIPS  $\downarrow$& FID  $\downarrow$&  LPIPS  $\downarrow$& FID  $\downarrow$&  LPIPS  $\downarrow$& FID  $\downarrow$\\
\midrule
DAE& {{0.047}}& 41.48& {{0.058}}& 41.52&  0.073& 41.55 \\
LDM& {{0.155}}& 22.49& {{0.352}}& 28.27&  0.666& 43.81 \\
DiffBIR Stage 1& 0.059& {{14.51}}&  0.334& 37.48&  0.426& 47.11 \\
DiffBIR (Full)& {\bf 0.024}& 13.52& 0.354& 29.80& 0.545& 46.67\\
LMPPM & 0.042& {18.68}& 0.058& 22.72& 0.063& 26.51 \\
LMPPM + DiffBIR Stage 2& {0.039}& {\bf{17.88}}& {\bf 0.055}& {\bf 18.49}& {\bf 0.059}& {\bf 25.49}\\
\midrule
\midrule
& \multicolumn{2}{c|}{Scribble 5}
& \multicolumn{2}{c|}{Nose Mask}
& \multicolumn{2}{c}{Left Eye Mask}\\
& LPIPS  $\downarrow$& FID  $\downarrow$&  LPIPS  $\downarrow$& FID  $\downarrow$&  LPIPS  $\downarrow$& FID  $\downarrow$\\
\midrule
DAE& {{0.069}}& 41.53& {{0.076}}& 41.49&  {\bf 0.086}& 41.48 \\
LDM& {{0.107}}& 37.87& {{0.071}}& 48.51&  {0.096}& 40.80 \\
DiffBIR Stage 1& 0.106& {{38.20}}&  0.081& 45.73&  0.099& 38.31 \\
DiffBIR (Full)& 0.102& 39.84& {\bf 0.050}& 48.69& {0.095}& 50.29\\
LMPPM & 0.078& {\bf 22.05}& 0.091& {\bf 17.15}& 0.137& {\bf 19.01} \\
LMPPM + DiffBIR Stage 2& {\bf 0.064}& {{24.24}}& {0.079}& {17.93}& {0.108}& {19.88}\\
 \bottomrule
 \end{tabular}
 \end{center}
\end{table}

\begin{table}
\begin{center}
 \caption{Quantitative results on CelebA-HQ-256}
 \label{tab:celebA}
\begin{tabular}{@{}l|ll|ll|ll@{}}
\toprule
& \multicolumn{2}{c|}{Noise 0.1}
& \multicolumn{2}{c|}{Noise 0.2}
& \multicolumn{2}{c}{Miss pixels 0.08}\\
& LPIPS  $\downarrow$& FID  $\downarrow$&  LPIPS  $\downarrow$& FID  $\downarrow$&  LPIPS  $\downarrow$& FID  $\downarrow$\\
\midrule
DAE& {\bf{0.018}}& 44.34& {{0.075}}& 44.43&  0.129& 43.82 \\
LDM& {{0.205}}& 27.24& {{0.512}}& 34.51&  0.736& 45.93 \\
DiffBIR Stage 1& 0.063& {{19.43}}&  0.459& 32.74&  0.436& 42.27 \\
DiffBIR (Full)& 0.047& {\bf 19.90}& 0.623& 34.05& 0.612& 44.75\\
LMPPM & 0.079& {24.72}& 0.086& 25.24& 0.107& {\bf 28.36} \\
LMPPM + DiffBIR Stage 2& {0.194}& {{24.91}}& {\bf 0.072}& {\bf 25.15}& {\bf 0.098}& {28.73}\\
\midrule
\midrule
& \multicolumn{2}{c|}{Scribble 5}
& \multicolumn{2}{c|}{Nose Mask}
& \multicolumn{2}{c}{Left Eye Mask}\\
& LPIPS  $\downarrow$& FID  $\downarrow$&  LPIPS  $\downarrow$& FID  $\downarrow$&  LPIPS  $\downarrow$& FID  $\downarrow$\\
\midrule
DAE& {{0.162}}& 44.27& {{0.099}}& 44.32&  0.122& 44.35 \\
LDM& {{0.192}}& 39.94& {{0.095}}& 46.13&  0.107& 40.97 \\
DiffBIR Stage 1& {\bf 0.159}& {{36.18}}&  {\bf 0.069}& 45.96&  {\bf 0.086}& 38.72 \\
DiffBIR (Full)& 0.174& 39.37& 0.082& 48.76& 0.104& 42.38\\
LMPPM & 0.193& {26.44}& 0.158& 26.33& 0.187& 26.19 \\
LMPPM + DiffBIR Stage 2& {0.163}& {\bf{26.35}}& {0.126}& {\bf 25.81}& {0.149}& {\bf 25.78}\\
 \bottomrule
 \end{tabular}
 \end{center}
\end{table}

\begin{figure}
    \centering
   \includegraphics[width=0.4\linewidth]{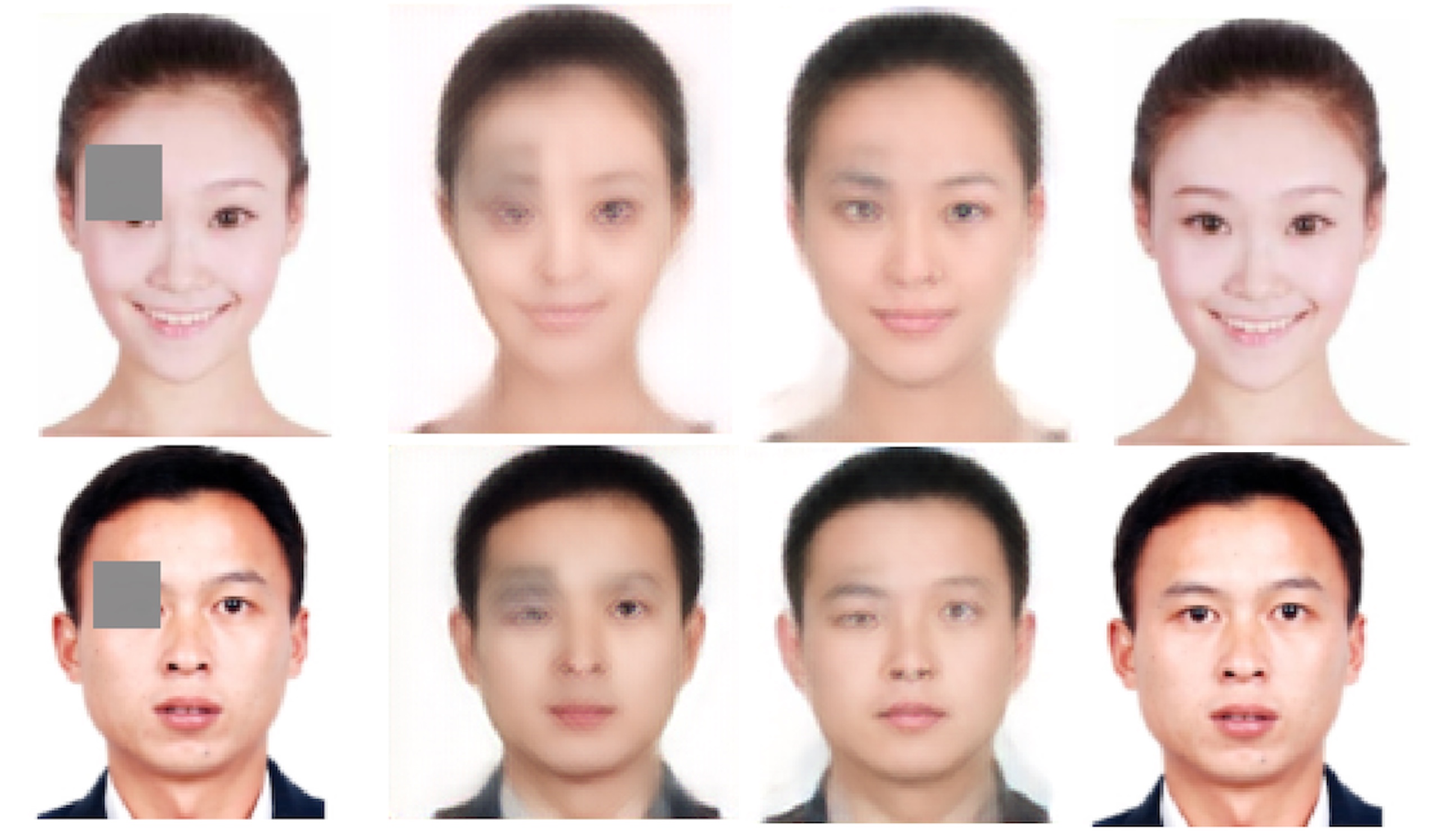}
   \caption{Ablation Eye Masking. From left to right: degraded, LMPPM with no ID loss, LMPPM, original.}
    \label{fig:ablation_noid}
\end{figure}

\begin{figure}
    \centering
   \includegraphics[width=0.4\linewidth]{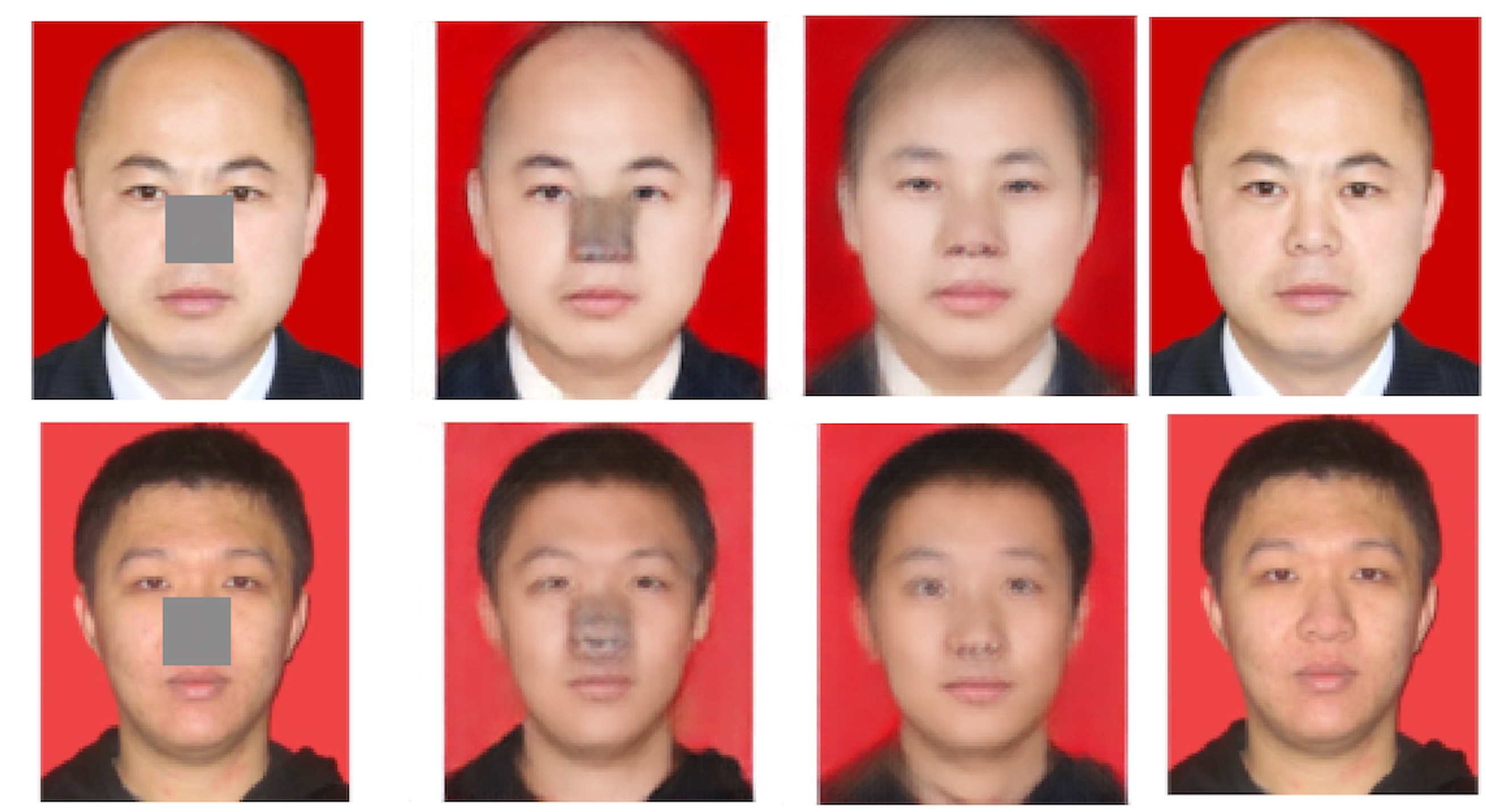}
   \caption{Ablation nose Masking. From left to right: degraded, LMPPM $\beta=0$, LMPPM, original.}
    \label{fig:ablation_beta0}
\end{figure}

\begin{figure}
    \centering
   \includegraphics[width=0.4\linewidth]{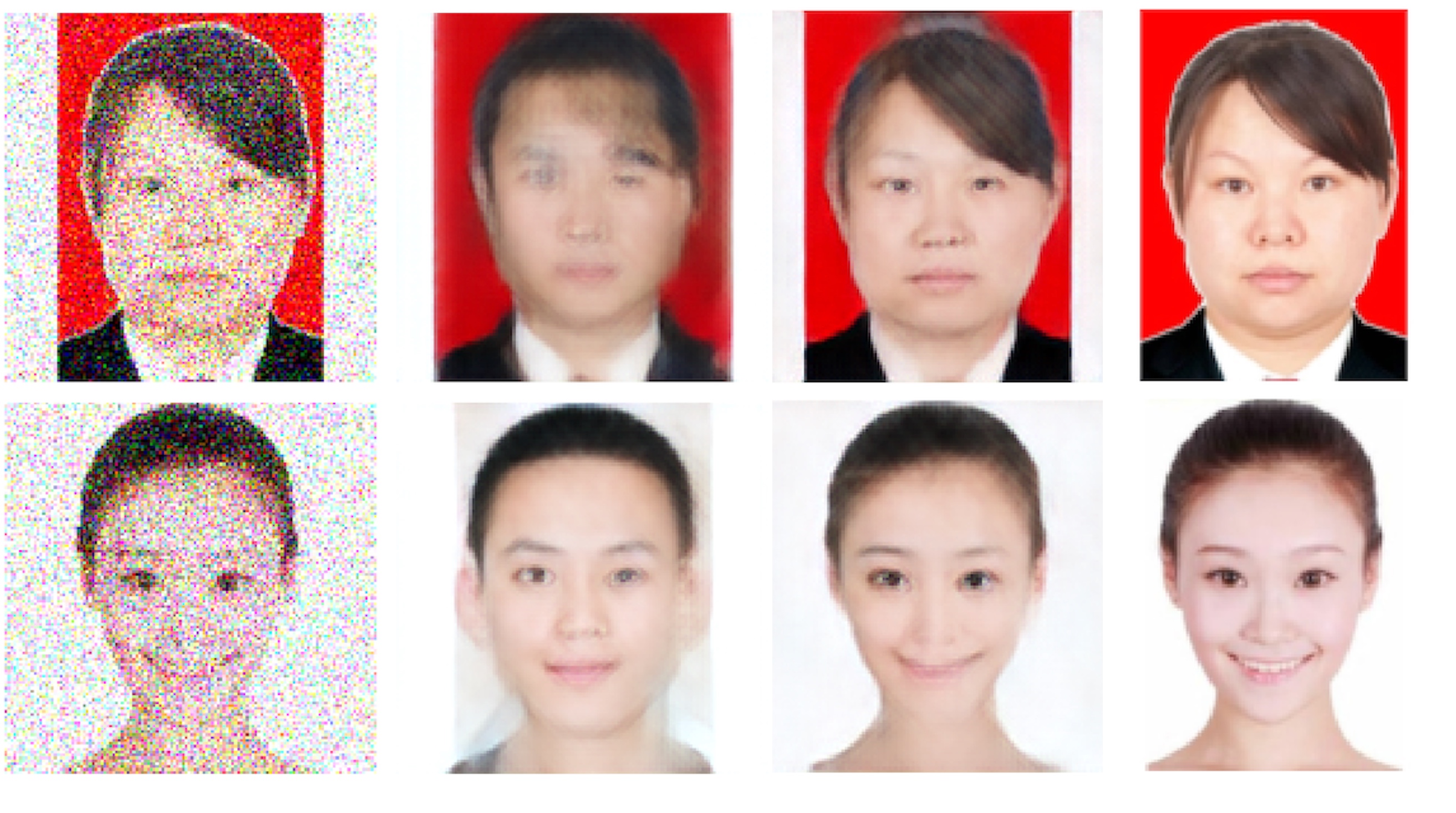}
   \caption{Ablation noise, $\sigma=0.2$. From left to right: degraded, LMPPM $\alpha=0$, LMPPM, original.}
    \label{fig:ablation_alpha0}
\end{figure}
\subsection{Ablation Study}
We further conducted several ablation studies, summarized in Table~\ref{tab:ablation}. First, we removed the ID loss from~\eqref{eq:main_loss_latent}. As shown in Fig.~\ref{fig:ablation_noid} and in the first column of Table~\ref{tab:ablation}, the full LMPPM loss produces better results. Next, we set $\beta=0$ for the nose-masking experiment. As shown in Fig.~\ref{fig:ablation_beta0}, the algorithm failed to complete the missing nose, since the $\bar{z}$ term was eliminated. Finally, we examined the case where $\alpha=0$. In this case, the algorithm relied entirely on $\bar{z}$ and generated a new face that was substantially different from the original. For example, in the bottom row of Fig.~\ref{fig:ablation_alpha0}, a female face was reconstructed as a male face. Note that the use of $\bar{z}$ assumes that the dataset used is dense enough such that any new degradation is influenced with $\bar{z}$ by images that are close enough. Otherwise $\bar{z}$ can introduce artifacts from images that are not similar enough to the degraded image.

\begin{table}
\begin{center}
 \caption{Ablation Study}
 \label{tab:ablation}
\begin{tabular}{@{}l|ll|ll|ll@{}}
\toprule
& \multicolumn{2}{c|}{No ID loss }
& \multicolumn{2}{c|}{$\beta=0$}
& \multicolumn{2}{c}{$\alpha=0$}\\
& LPIPS  $\downarrow$& FID  $\downarrow$&  LPIPS  $\downarrow$& FID  $\downarrow$&  LPIPS  $\downarrow$& FID  $\downarrow$\\
\midrule
Ablation& {{0.144}}& 22.20& {\bf{0.066}}& 19.78&  0.131& 24.13 \\
LMPPM& {\bf{0.137}}& {\bf 19.01}& {{0.091}}& {\bf 17.15}&  {\bf0.058}& {\bf 22.72} \\
 \bottomrule
 \end{tabular}
 \end{center}
\end{table}

\section{Summary and Conclusions}
This work develops a geometric framework for image restoration based on the \emph{Manifold Hypothesis}, providing a unified perspective on image restoration methods. Our key idea is to combine a learned distance function to the manifold of clean images with a kernel-like approximation of the probability density on the manifold, thereby integrating geometric structure and probability. The resulting score-induced vector field guides corrupted samples toward regions of the ambient space corresponding to both the structure and the distribution of clean images.

A central practical advantage of the proposed approach is its compact architecture and data-efficient training. The networks defining the manifold, distance function, and probability approximation are compact, with fewer than 6M parameters in total, and can be trained using only a few thousand clean samples. Moreover, the models are trained exclusively with Gaussian noise, yet are evaluated at inference time under a substantially broader range of degradation types such as masking and scribbles. This provides a challenging test of whether the learned manifold prior captures properties of clean images that generalize beyond the specific degradation used during training. Our experiments demonstrate promising restoration performance under Gaussian noise as well as unseen degradations, including scribbles and masking, across multiple datasets.

The iterative nature of our restoration procedure provides another practical advantage: the number of restoration steps does not need to be estimated in advance. Rather than specifying a fixed number of iterations, we monitor the restoration process and stop when the relevant probability measure reaches a plateau. This adaptive stopping criterion allows the procedure to terminate naturally once further iterations no longer provide meaningful progress.

Operating in latent space is also important to the effectiveness of the method. The reduced dimensionality facilitates more accurate learning of the distance function and probability density, while preserving the geometric structure relevant to image restoration. Comparisons with established baselines, including denoising autoencoders, latent diffusion models, and DiffBIR on facial datasets, demonstrate the potential of the proposed approach, particularly under severe and diverse distortions.
This work is currently limited to additive degradations, including Gaussian noise and other additive non-kernel based corruptions. An important direction for future work is to extend the proposed framework to more com-lex degradation models, such as blur, spatially varying noise, and combinations of these types of kernel based degradation with other types. These types of degradation are harder to handle with our current algorithm because of the autoencoder's limited ability to separate clean images from blurry ones. This manifests itself in the well-known tendency of autoencoders to generate blurry images. From our geometric perspective, blurry images lie inside the manifold of clean images, which is the main cause of the difficulty. We are currently investigating these issues.

\section*{Acknowledgments}
We gratefully acknowledge Prof. Yoel Shkolnisky and Prof. Haim Avron for fruitful discussions in the early stages of the project, held as part of the departmental research group. We also thank Prof. Yoel Shkolnisky and Dr. Dmitry Batenkov for generously providing unrestricted access to their computational infrastructure, including GPU servers, which was essential for this research.


\clearpage
\appendix
\section{DDPM}
\label{app:ddpm}
Here we revise the Denoising Diffusion Probabilistic Model (DDPM) procedure for train and inference.
\begin{algorithm}[H]
\caption{DDPM Training}
\begin{algorithmic}[1]
\State \textbf{Precompute noise schedule:}
\State $\beta_t$ (linear schedule from $0.0001$ to $0.02$)
\State $\alpha_t = 1 - \beta_t$
\State $\bar{\alpha}_t = \prod_{i=1}^{t} \alpha_i$
\For{\textbf{each batch}}
    \State Sample timestep $t \sim \text{Uniform}(0, T-1)$
    \State Sample noise $\epsilon \sim \mathcal{N}(0, I)$
    \State \textbf{Forward diffusion:} 
    \[
    x_t = \sqrt{\bar{\alpha}_t}\, x_0 + \sqrt{1 - \bar{\alpha}_t}\, \epsilon
    \]
    \State Predict noise: $\epsilon_{\text{pred}} = \text{model}(x_t,\, t/T)$
    \State Compute loss:
    \[
    L = \|\epsilon - \epsilon_{\text{pred}}\|^2
    \]
    \State Backpropagate and update parameters
\EndFor
\end{algorithmic}
\label{alg:diff_train}
\end{algorithm}
\begin{algorithm}[H]
\caption{DDPM Inference (Stochastic, 1000 steps)}
\begin{algorithmic}[1]
\State Initialize $x_T \sim \mathcal{N}(0, I)$
\For{$t = T-1$ \textbf{down to} $0$}
    \State Predict noise: $\epsilon_{\text{pred}} = \text{model}(x_t,\, t/T)$
    \State \textbf{Denoise:}
    \[
    x_t = \frac{x_t - \text{coef}_t \, \epsilon_{\text{pred}}}{\sqrt{\alpha_t}}
    \]
    \If{$t > 0$}
        \State Sample $z \sim \mathcal{N}(0, I)$
        \State Add noise: $x_t = x_t + \sigma_t z$
    \EndIf
\EndFor
\State \Return $x_0$
\end{algorithmic}
\label{alg:diff_inference}
\end{algorithm}

\section{Detailed Theory}
\label{app:theory}

\subsection{{Manifold Likelihood Derivation}}
\label{app:non_uniform}

Here we derive Eq.~\eqref{eq:kernel}. Let $x$ denote a general point in the ambient space, and let $x'$ be a clean data point. We denote the probability density of clean images by $P_c(x')$. Then, by~\eqref{cond-prob},
\begin{equation*}
    \begin{aligned}
    P_{{\text{non-u}}}(x;\sigma) &= \int_\mi P(x,x')dx' = \int_{\mathbb{R}^D} P_\sigma(x\mid  x') P_c(x')dx'   \\
    &=\int_{\mathbb{R}^D} P_\sigma(x\mid  x')\left(\int_\mi P(G(z))\delta(x'-G(z))dV_\mi\right)dx'\\
    &= \int_\mi P_\sigma(x\mid x'=G(z))P(G(z))\underbrace{\sqrt{g}dz}_{dV_\mi}=\int_{\mathbb{R}^d} P_\sigma(x\mid G(z))P(z)dz. 
    \end{aligned}
\end{equation*}
Note that $P_{{\text{non-u}}}$ is the combination of $P(x;\sigma)$ and the manifold hypothesis. Here $dV_\mi=\sqrt{g}dz$ is the manifold’s volume element, where $g=\det G$ and $G_{\mu\nu}=\sum_{i=1}^D J^i_\mu J^i_\nu$ is the induced metric, with the Jacobian of the embedding map given by $J^i_\mu=\partial G^i(z)/\partial z_\mu$. In the last equality, we use the identity $P(G(z))=P(z)(\sqrt{g})^{-1}$.

\subsection{Score Function}
\label{app:score}
Recall the score function
$$
s(x)=\nabla_x \log {P_{{\text{non-u}}}}(x)\approx \nabla_x\log \hat{P}_{{\text{non-u}}}(x) =:\hat{s}(x).
$$
Direct computation results in 
\begin{equation}\label{eq:s_k_hat_bis}
\hat{s}(x) = -\frac{1}{2\sigma^2}\left( x-\bar{G}(x)\right), 
\end{equation}
where $\bar{G}(x)=\sum_{\alpha\in S} \bar{G}_\alpha(x)$, and
\begin{equation}
\bar{G}_\alpha(x) =  \frac{1}{\hat{P}_{{\text{non-u}}}(x)Q_{\ker}}\int\left[ G(z) P(x\mid G(z))\exp\left({-\frac{\|z-z_\alpha\|^2}{2\sigma_{\text{ker}}^2}}\right)\right]dz.
\label{eq:app_g_bar}
\end{equation}   
{We approximate the integral over $z$
in \eqref{eq:app_g_bar} for $\bar{G}_\alpha(x)$ by Monte Carlo sampling from the normal distribution centered at the training point $z_\alpha$. Specifically, we estimate the mean by averaging over $n$ samples. }
\begin{equation}
    \int\left[ G(z) P(x\mid G(z))\exp\left({-\frac{\|z-z_\alpha\|^2}{2\sigma_{\ker}^2}}\right)\right]dz\approx \frac{1}{Q_\sigma}\frac{1}{n}\sum_{z_i\in \mathcal{N}(z_\alpha,\sigma_{\ker}^2)}
    G(z_i)\exp\left({-\frac{\|{x}-G(z_i)\|^2}{2\sigma^2}}\right),
    \label{eq:app_cond}
\end{equation}
where $\alpha$ denotes an index in the training set. The calculation of $\bar{G}_\alpha(x)$ requires evaluating $\hat{P}_{\text{non-u}}(x)$ in the denominator. In particular, by~\eqref{eq:pnonu}
\begin{equation}
    \hat{P}_{\text{non-u}}=\frac{1}{Q_{\text{ker}}}\int\left[ P(x\mid G(z))\exp\left({-\frac{\|z-z_\alpha\|^2}{2\sigma_{\text{ker}}^2}}\right)\right]dz\approx \frac{1}{Q_{\text{ker}}Q_\sigma}\frac{1}{n}\sum_{z\in \mathcal{N}(z_\alpha,\sigma_{\ker}^2)}
    \exp\left({-\frac{\|{x}-G(z)\|^2}{2\sigma^2}}\right).
\label{eq:app_pnonu}
\end{equation}
{Substituting~\eqref{eq:app_cond} and ~\eqref{eq:app_pnonu} in~\eqref{eq:app_g_bar} yields
\begin{equation}
\bar{G}_\alpha(x)\approx \frac{\sum_{z\in \mathcal{N}(z_\alpha,\sigma_{\ker}^2)}
    G(z)\exp\left(-{\|{x}-G(z)\|^2/2\sigma^2}\right)}{\sum_{z\in \mathcal{N}(z_\alpha,\sigma_{\ker}^2)}
    \exp\left({-\|{x}-G(z)\|^2/2\sigma^2}\right)}.
\end{equation}
}
Note that all constant factors $Q_\sigma,\ Q_{\text{ker}}$ and $\frac{1}{n}$, are canceled between the numerator and the denominator. 

\subsection{The Tweedie Formula}
\label{app:tweedie}
The ``flow'' equations \eqref{eq:x_follow_D} and \eqref{eq:x_follow_D_G_avg} are the Tweedie formulas \cite{Efron2011} for the corresponding probability functions where \eqref{eq:x_follow_D} follows 
\begin{equation}
\label{distance-prob-bis}
P_d({x}) = \frac{1}{Q_d}\exp\left({-\frac{\alpha}{2}{\di^2_\mi({x})}}\right), 
\end{equation}
and~\eqref{eq:x_follow_D_G_avg} follows
\begin{equation}
\label{combine-prob}
P({x}) = \frac{1}{Z}P_d({x})\sum_{\alpha\in S}\int_{\mathbb{R}^d} \exp\left({-\frac{\beta}{2}{\|{x}-{G}(z)\|^2}}\right)\exp\left({-\frac{\|z-z_\alpha\|^2}{2\sigma_{\text{ker}}^2}}\right)dz, 
\end{equation}
where $Z$ is a normalization factor. One can easily verify from \eqref{eq:x_follow_D} that we decrease the distance to the manifold along the flow. Indeed, 
\begin{equation*}
 \begin{aligned}
    \di_\mi(x^{n+1}) &= \di_\mi(x^n - \epsilon \di_\mi(x^n)\nabla_x \di_\mi(x^n))\cr 
    &= \di_\mi(x^n) - \epsilon \nabla \di_\mi(x^n)\cdot \di_\mi(x^n)\nabla_x \di_\mi(x^n) + O(\epsilon^2) = (1-\epsilon)\di_\mi(x^n) + O(\epsilon^2),
    \end{aligned}
\end{equation*}
where we used the fact that the distance function is a solution of the Eikonal equation $\|\nabla_x \di_\mi(x)\|^2 = 1$.

\section{Detailed Experimental Setup}
\label{app:exp_setup}

\subsection{Notation and Abbreviations}
In the following tables we summarize the abbreviations and datasets through the paper.
\begin{table}[ht!]
\centering
\caption{Glossary of abbreviations and terms used throughout the paper}
\begin{tabular}{ll}
\hline
\textbf{Term} & \textbf{Definition} \\
\hline
DAE & Denoising Autoencoder \\
MPPM& Manifold Projection and Propagation Method (our proposed approach) \\
LMPPM & Latent Manifold Projection and Propagation Method (our proposed approach) \\
LDM & Latent Diffusion Model \\
SSIM & Structural Similarity Index Measure \\
GN & Group Normalization \\
\hline
\end{tabular}
\end{table}
\begin{table}[ht!]
\centering
\caption{Summary of experimental datasets used for evaluating restoration performance}
\begin{tabular}{ll}
\hline
\textbf{Dataset} & \textbf{Description} \\
\hline
MNIST & 60,000 training/10,000 test grayscale images ($28 \times 28$ pixels) \\
SCUT-FBP5500 & 5,500 facial images with beauty scores (resized to $120 \times 120$) \\
CelebA-HQ-256& 30,000 high-quality celebrity face images ($ 256\times 256$)\\
\hline
\end{tabular}
\end{table}
\subsection{Degradations}
\paragraph{Degradation Parameters}
We apply six degradation types to simulate real-world image corruption scenarios. Each degradation is applied at two severity levels (intermediate, and severe) to test the robustness of restoration methods.

\paragraph{Degradation Methods} Brief descriptions of each degradation type:
\begin{itemize}
\item \textbf{Gaussian noise}: Additive zero-mean Gaussian noise that simulates sensor noise or transmission errors.
\item \textbf{Elastic deformation}: Non-rigid distortions implemented using \\
\texttt{torchvision.transform.ElasticTransform($\alpha=34$,$\sigma$)} that simulate warping effects.
\item \textbf{Super-resolution}: Downsampling followed by upsampling to original resolution, simulating reconstruction from low-resolution data.
\item \textbf{Missing Pixels}: Set black patches with some coverage portion
\item \textbf {Scribbles}: Add $n$ random scribbles with random colors
\item \textbf {Partial Masking}: mask part of the face area e.g. the nose or an eye.
\end{itemize}

\subsection{Model Architectures}

We implemented three main architectures across all experiments, with design choices tailored to each dataset's complexity.

\paragraph{Synthetic Data Model}
Synthetic data for MPPM experiments use MLP-based networks with a latent dimension of 1, selected based on the low intrinsic dimensionality of this manifold:

\begin{table}[h]
\centering
\caption{Network architectures for synthetic data experiments. All models use fully-connected layers}
\begin{tabular}{ll}
\hline
\textbf{Component} & \textbf{Architecture} \\
\hline
Encoder & $3 \to 64 \to 64  \to 1$ with ReLU  \\
Decoder & $1 \to 64 \to 64  \to 3$ with ReLU \\
Distance Network & $3 \to 16 \to 8 \to 4 \to 1$ with ReLU \\
\hline
\end{tabular}
\end{table}

\paragraph{MNIST Models}
MNIST experiments use CNN-based models with latent dimension 18, chosen to capture the variability among handwritten digits while promoting compact representations.

\begin{table}[h]
\caption{Network architectures for MNIST experiments}
\begin{tabular}{ll}
\hline
\textbf{Component} & \textbf{Architecture} \\
\hline
Encoder & $\text{Conv2d}(1\to 32\to 64, \text{kernel}=3, \text{stride}=2) \to \text{Flatten} \to \text{Linear}(64\times 7\times 7\to 18)$ \\
Decoder & $\text{Linear}(18 \to 64\times 7\times 7) \to \text{Reshape} \to \text{ConvTranspose2d}(64\to 32\to 1) \to \text{Sigmoid}$ \\
Distance Network & $18 \to 100 \to 50 \to 20 \to 1$ with ReLU, dropout=0.2 \\
\hline
\end{tabular}
\end{table}

\begin{figure}
\includegraphics[width=\linewidth]{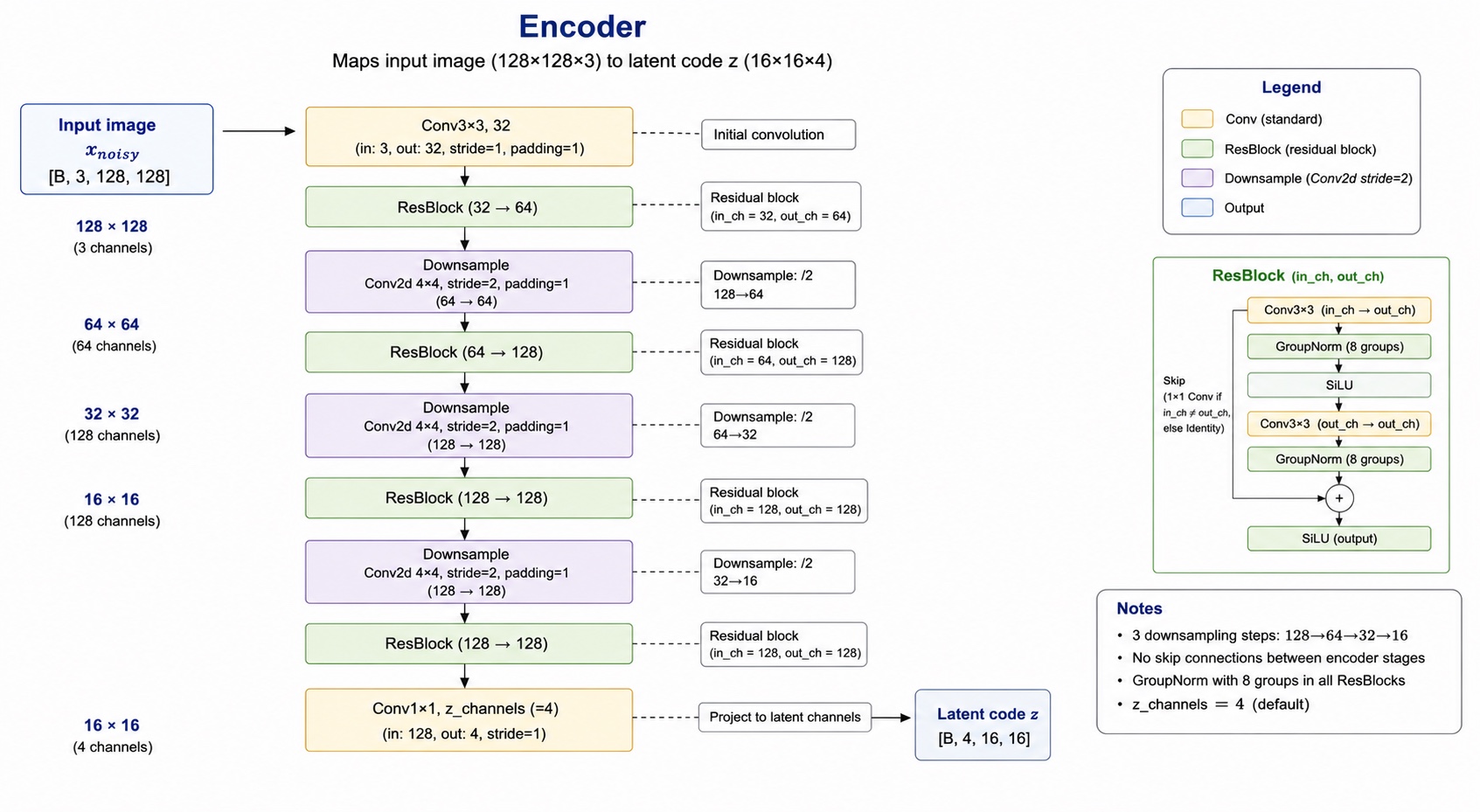}
\caption{Architecture of Encoder $F$}
\label{fig:encoder}
\end{figure}

\begin{figure}
\includegraphics[width=\linewidth]{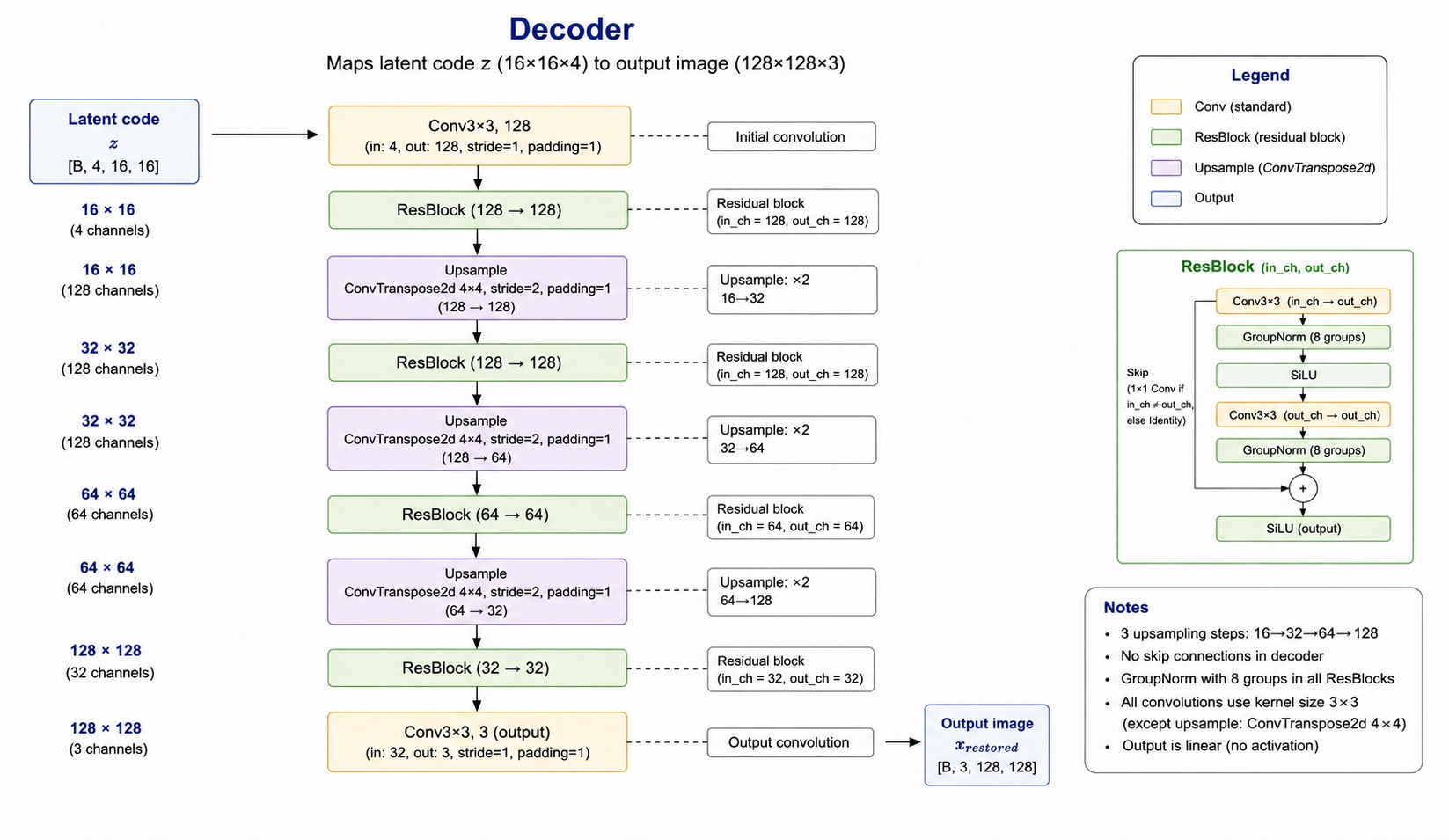}
\caption{Architecture of Decoder $G$}
\label{fig:decoder}
\end{figure}

\begin{figure}
\includegraphics[width=\linewidth]{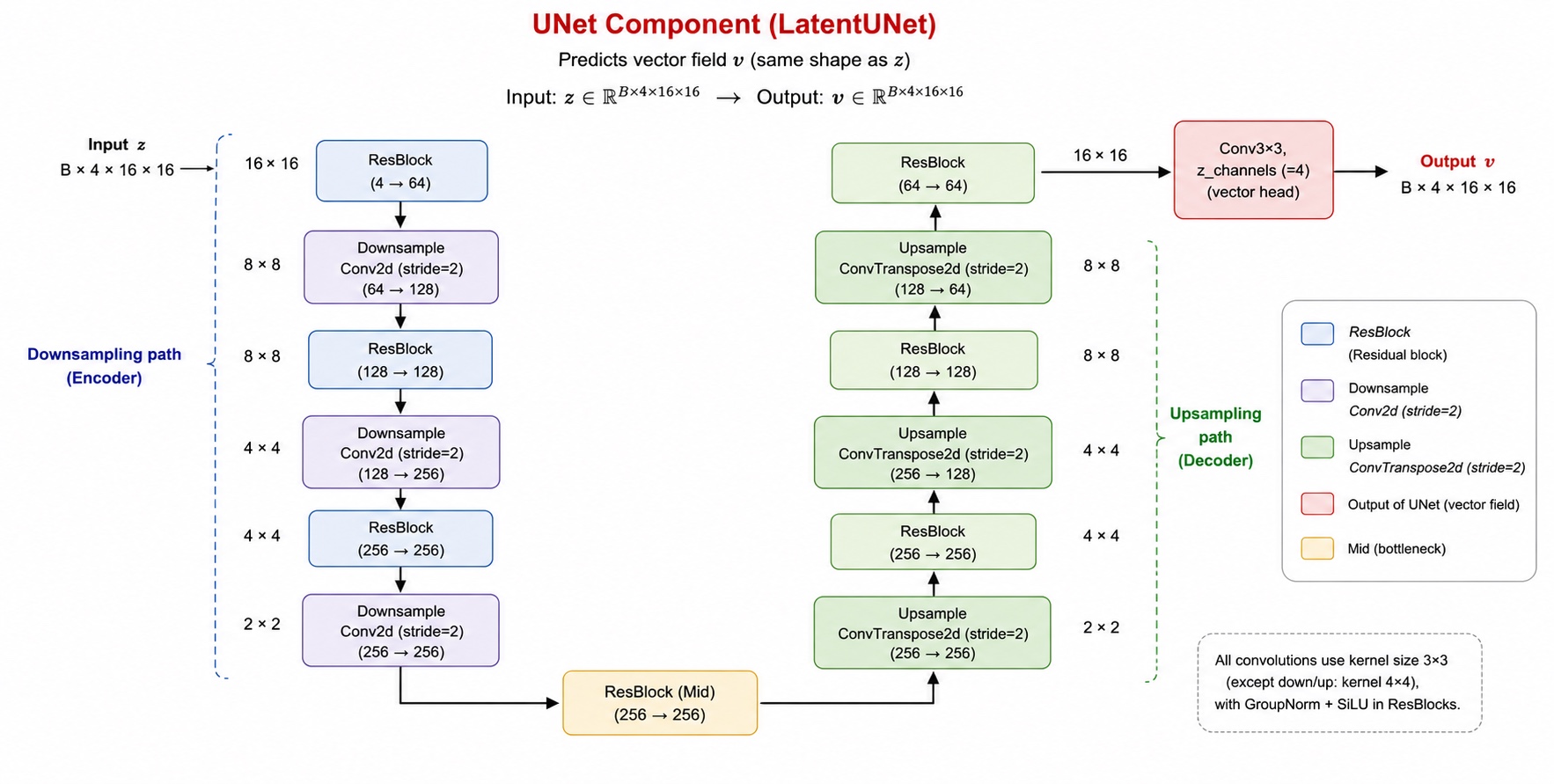}
\caption{Architecture of Direction Field $V$ implemented as Unet}
\label{fig:v_direction}
\end{figure}

\begin{figure}
\includegraphics[width=\linewidth]{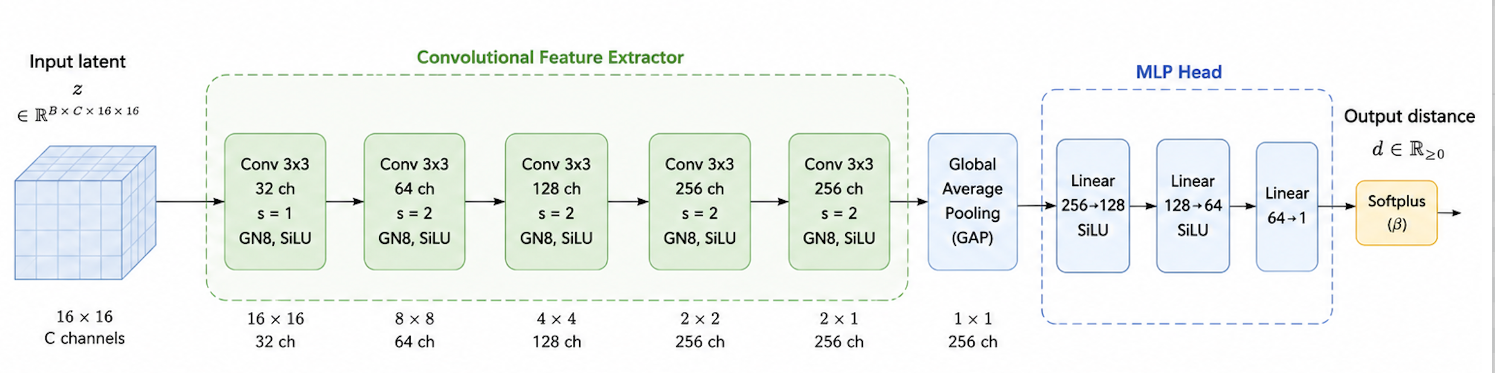}
\caption{Architecture of Distance network $\di_S(z)$}
\label{fig:dist}
\end{figure}

\paragraph{SCUT-FBP5500 and CelebA-HQ-256 Models}
Facial image experiments employ a latent dimension of 1024, which accommodates the higher complexity of facial features while enabling detailed reconstruction. The architectures of the encoder $F$, decoder $G$, direction field $V$ and distance function $\di_S$ are shown in figures~\ref{fig:encoder}-\ref{fig:dist}.

%

\section{Training and Evaluation}
\label{app:train_and_eval}

\begin{table}[h!]
\caption{MPPM training and inference parameters for synthetic data}
\centering
\begin{tabular}{ll}
\hline
\textbf{Parameter} & \textbf{Value} \\
\hline
Optimizer & Adam ($\beta_1=0.9$, $\beta_2=0.999$) \\
Learning rates & AE: $1\times 10^{-3}$, Distance network: $1\times 10^{-3}$ \\
Weight decay & $1\times 10^{-4}$ \\
Batch size & 550 \\
Training epochs & 1500 \\
Loss function & Composite loss (Equation \ref{eq:main_loss_ambient}) \\
Early stopping & Patience: 100 \\
\hline
$\alpha$ (distance gradient step) & 0.15 \\
$\beta$ (kernel averaging weight) & 0.1 \\
Convergence tolerance & 0.005 \\
Maximum iterations & 60 \\
\hline
\end{tabular}
\end{table}

\begin{table}[h!]
\centering
\caption{LMPPM training parameters across all experiments, determined through preliminary grid search, diffusion steps are define the number of steps in algorithm \ref{alg:lmppm} }
\begin{tabular}{ll}
\hline
\textbf{Parameter} & \textbf{Value} \\
\hline
Optimizer & MIST: Adam ($\beta_1=0.9$, $\beta_2=0.999$), Faces: Adamw (weight decay=$0.001$ \\
Learning rates & MNIST: AE: $1\times 10^{-3}$, Distance network: $1\times 10^{-3}$ (MNIST $1\times 10^{-5})$, LDM: $1\times 10^{-3}$ \\
& Faces: $3.0\times 10^{-4}$\\
Batch size & MNIST: 128, Faces: 32 \\
Training epochs & MNIST: 200, Faces: 100 \\
Loss functions & DAE: L2, LDM: MSE, LMPPM: Composite loss~\eqref{eq:main_loss_latent} \\
Early stopping & Patience: 10 epochs \\
Diffusion steps & MNIST: 2000, Faces: 2000 \\
\hline
\end{tabular}
\end{table}
\newpage
\vskip 0.2in
\bibliographystyle{unsrt}
\bibliography{ref}
\end{document}